\documentclass[10pt,journal,compsoc]{IEEEtran}
\usepackage{graphicx}
\usepackage{amsmath}
\usepackage{amssymb}
\usepackage{amsfonts}
\usepackage{array}
\usepackage{pgf}
\usepackage{pgfplots}
\usepackage{tikz}
\usepackage{ragged2e}
\usepackage{url}
\usepackage{algorithm}
\usepackage{algorithmic}
\usetikzlibrary{calc,patterns,angles,quotes}
\usepackage{graphicx}
\ifCLASSOPTIONcompsoc
\usepackage[caption=false, font=normalsize, labelfont=sf, textfont=sf]{subfig}
\else
\usepackage[caption=false, font=footnotesize]{subfig}
\fi
\ifCLASSOPTIONcompsoc
\usepackage[nocompress]{cite}
\else
\usepackage{cite}
\fi
\usepackage[capitalise]{cleveref}
\DeclareMathOperator*{\E}{\mathbb{E}}

\DeclareMathOperator{\I}{\mathbb{I}}
\renewcommand{\raggedright}{\leftskip=0pt \rightskip=0pt plus 0cm}
\begin{document}
	
\author{Yang~Li, Yaqiang~Yao,
		\IEEEcompsocitemizethanks{\IEEEcompsocthanksitem The authors are with School of Computer
			Science and Technology, University of Science and Technology of China
			(USTC), Hefei, Anhui 230027, China. E-mail: \{csly, yaoyaq\}@mail.ustc.edu.cn }}

\title{Bayesian Optimization with Directionally Constrained Search}

\IEEEtitleabstractindextext{
\begin{abstract}
\raggedright{
Bayesian optimization offers a flexible framework to optimize an
	objective function that is expensive to be evaluated. A Bayesian optimizer
	iteratively queries the function values on its carefully selected points.
	Subsequently, it makes a sensible recommendation about where the optimum locates
	based on its accumulated knowledge. This procedure usually demands a long
	execution time. In practice, however, there often exists a computational budget
	or an evaluation limitation allocated to an optimizer, due to the resource
	scarcity. This constraint demands an optimizer to be aware of its remaining
	budget and able to spend it wisely, in order to return as better a point as
	possible. In this paper, we propose a Bayesian optimization approach in this
	evaluation-limited scenario. Our approach is based on constraining searching
	directions so as to dedicate the model capability to the most promising area. It
	could be viewed as a combination of local and global searching policies, which
	aims at reducing inefficient exploration in the local searching areas, thus
	making a searching policy more efficient. Experimental studies are conducted on
	both synthetic and real-world applications. The results demonstrate the superior
	performance of our newly proposed approach in searching for the optimum within a
	prescribed evaluation budget. }
\end{abstract}

}

\maketitle
\IEEEdisplaynontitleabstractindextext
\IEEEpeerreviewmaketitle

\IEEEraisesectionheading{\section{Introduction}}
\label{sec:intro}

\IEEEPARstart{M}{achine} learning is rarely optimization-free. Traditional
numerical optimizers make use of the structural properties of the problems like
convexity, submodularity etc. to accelerate their convergence rates. However,
there are also a number of problems that traditional optimizers are
insufficient. A typical example is tuning the hyper-parameters for some machine
learning algorithms such that the validation error is minimized. In such a
problem, there is no closed-form for the objective function, not to mention its
structural properties. The hyper-parameters, which are considered as nuisances,
need to be repeatedly changed and compared. When the objective function is
computationally prohibitive to be executed repeatedly, traditional numerical
optimizers become insufficient.  Bayesian Optimization (BO) \cite{Mockus1975}
offers a choice for such \emph{black-box} problems, and has achieved a decent
performance on a number of challenging benchmark functions \cite{Jones2001}.

For continuous objective functions, Bayesian optimization typically works by
presuming the uncertainty of not-yet observed function values follows a
multivariate Gaussian distribution. As new observations are made, it updates the
belief about global minimizer and infers the next point that should be queried.
For instance, in the problem of tuning the hyper-parameters of a machine
learning algorithm, BO treats it as a global searching on an expensive,
black-box function plane, whose derivatives are unaccessible. The algorithm of
interest is invoked under different hyper-parameter settings. As more
experimental results are observed, BO recommends a setting with high probability that
will achieve a high performance based on its previous trails and accumulated knowledge. 

BO is an iterative procedure that successively queries function values and updates
a posterior distribution for not-yet evaluated points. Its performance depends critically on the strategy of selecting the
next query point since this step enables it to actively learn the structure of the black-box
function. This strategy is explicitly expressed via an \emph{acquisition function},
which measures how much improvement or utility a not-yet-evaluated point would bring if it were
selected.

Acquisition function distinguishes different Bayesian optimization techniques. A
list of most popular ones include: Expected Improvement (EI) \cite{Mockus1975}
over the present best result; Upper Confidence Bound (UCB) \cite{Srinivas2010};
Probability of Improvement (PoI) \cite{Kushner1964}, which introduces a
parameter $ \kappa $ to trade-off exploring against exploiting. By exploitation,
we mean that an algorithm always query a point that is most likely to be the
optimum from its knowledge, without considering the ones that are associated
with great uncertainties. Exploration, on the other hand, aims at reducing
uncertainties on not-evaluated points. EI and UCB have demonstrated their
efficiencies on finding the global optimum of multi-modal black-box functions
\cite{Srinivas2012,Bull2011}. Other particularly successful designs are Entropy
Search (ES) \cite{Hennig2012} and Predictive Entropy Search (PES)
\cite{Hernandez-Lobato2015,hernandez-lobato_predictive_2016}, both of which
involve maximizing the mutual information between a query point and a guess on
the global minimizer.

\begin{figure}[t]
	\centering \includegraphics[width=0.8\linewidth]{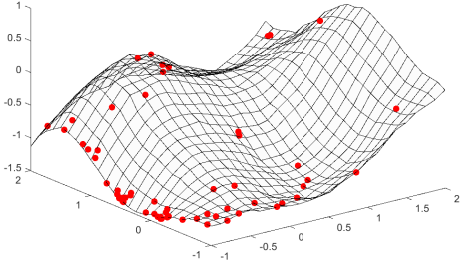}
	\caption{ The query points selected by Expected Improvement (EI) on a
		non-convex surface. Although some queries are actually near the global
		minimizer, there is still a large among of trails jiggling around,
		bring no benefits but leading to a waste of evaluation budget. This oscillating
		behavior near the minimizer is what we want to avoid. } \label{fig:ep-no-anneal}
\end{figure}

All the above approaches are effective in the sense they will ultimately locate
the global optimum, provided unlimited evaluation time. However, due to resource
scarcity such as high-performance computers, evaluation time and even patience
of practitioners, most BOs are executed within a given permissible number of
iterations or \emph{evaluation budget}. As a result, those approaches still keep
a high level of exploring even when they are near the maximum iteration limit.
This leads to a resource waste and excessively oscillating behavior on both the global and local scales.
An example in \cref{fig:ep-no-anneal} elucidates how this affects the behavior
of a BO algorithm.

To reduce the oscillating behavior often needs assistance from the structural
properties of the objective function, in particular the gradients. However, as we are
dealing with a black-box problem, we cannot assume that its gradient
information is accessible even locally. An immediate intuition is that,
along with the execution of an optimizer, a global profile or a surrogate model
about the objective function is made more and more accurate. We notice that,
while it is not possible to utilize the gradients, we may extract some
directional information from this constructed surrogate model, and encode our preference on
the searching directions into the acquisition function of a BO algorithm. 

In this work, we focus on the scenario where evaluation budget exists, and
propose an approach that imposes a directional constraint with increasing
effects on the searching policy. The whole procedure has several appealing
properties: (1) At the early steps of an optimization, an optimizer is allowed
to select the next query point freely by maximizing its acquisition function.
(2) However, since there is a strict limit on the number of iterations or
executions, it selects the next point to be evaluated from a direction that is
more likely to yield decrease. Besides that, we include another requirement for
this searching direction: it should span an acute angle with the last one on an
optimization path. This requirement is imposed for the prevention of jiggles.
(3) At the later stage especially near the end of iterations, this optimizer
gradually dedicate most of its probability into a local searching in the most
promising region as it becomes more rewarding. (4) This directional constraint
takes effects in the form of probability gradually over the execution of an
algorithm. Thus, our approach could be viewed as a novel combination of both
global and local searching policies.

We will focus our discussion on a typical acquisition function: Expectation
Improvement (EI), as it has been shown to be well-behaved and widely used
\cite{Bishop2006}. Different from the method UCB, it does not own any tuning
parameters. In addition, its simplicity on formulation is an appealing
characteristic on perception. But we will show that our newly proposed
constraint could easily be included into most common probabilistic acquisition
functions like EI, PoI, and UCB.

%

This paper is organized as follows. In \cref{sec:background}, we list Gaussian
process, Bayesian optimization, and expected improvement as a particular
example of acquisition functions. Section \ref{sec:method} introduces the
directional constraint and outlines the optimization procedure under the effects
from this newly proposed constraint. The experimental results and discussions
are then presented in \cref{sec:exp}, showing our approach is more efficient
than existing methods. Finally, section \ref{sec:conclusion} concludes the paper
and present future work, while Appendix lists the technical details that are
omitted in the text.

\section{Background}
\label{sec:background}

\subsection{Gaussian Process} 

A Gaussian Process (GP) is defined on a uncountable set of random variables. Any
selected variables from this set are subject to a joint Gaussian distribution. Formally,
data set $ \mathcal{D} $ is a collection of $ n $ points of dimension $ d
$, $ X = \{x_1,x_2,\cdots,x_n\} $. The corresponding real-valued function
values are denoted by $ f(X) = \{f({x}_1),f({x}_2),\cdots,f({x}_n) \} $. A
functional viewpoint treats GP as a distribution defined over functions
\cite{Rasmussen2006}, which compactly describes the uncertainty on $ f(\cdot ) $ via
a mean function $ \mu(\cdot) $ and a covariance kernel $ k(\cdot, \cdot) $. Some
key properties such as stationarity, smoothness etc. are specified through the
covariance function. For instance, a hyper-parameter $ \theta $ is routinely included in
the covariance function for the sake of versatility, which controls its kernel
width:
\begin{equation}
\label{eq:kernel}
	k(x_i,x_j) = exp\big( -\frac{1}{2\theta}|| x_i - x_j|| \big)
\end{equation}

Let $ x^\prime $ denotes an arbitrary query point. 
The GP tells us that the joint distribution of function values on the query point
and the collection of samples should satisfy:

\begin{equation*}
\left[
	\begin{matrix}
	f(X)\\ f(x^\prime)
	\end{matrix}\right] \sim \mathcal{N}
	\bigg(\left[
	\begin{matrix}
	\mu(X)\\ \mu(x^\prime)
	\end{matrix}\right]
	, \left[
	\begin{matrix}
	k(X,X) & k(X,x^\prime)\\
	k(x^\prime,X)& k(x^\prime,x^\prime)
	\end{matrix}\right]
	 \bigg)
\end{equation*}
where the operations should be understood as element-wisely on their matrix inputs. $
\mathcal{N}(\cdot) $ is a multi-variate Gaussian distribution. 


The posterior distribution on the query point $ x^\prime $ is calculated by standard Bayesian
rules. The distribution for the function value on $ x^\prime $, which is represented as $ f(x^\prime) \sim p(f(x^\prime)|x^\prime,X, f(X)) $, can be calculated as:
\begin{align*}
	{f}(x^\prime) &\sim p({f}(x^\prime)|X,f(X),x^\prime)  \\
	&= \mathbb{E}_{p({f}(X)|X)} p({f}(x^\prime)|X, {f}(X), x^\prime)\\
	&= \mathcal{N}(\mu(x^\prime), {\Sigma}(x^\prime))
\end{align*}
 where
\begin{align}
\label{update}
\begin{split}
\mu(x^\prime) &= \mu(x^\prime) + k(x^\prime,X)k(X,X)^{-1}(f(X) - {f}(X)) \\
\Sigma(x^\prime) &= k(x^\prime,x^\prime) - k(x^\prime,X)k(X,X)^{-1}k(X,x^\prime)
\end{split}
\end{align}

Being viewed as a machine learning algorithm, GP uses a measure of similarity to
make prediction on an unseen point. As opposed to a traditional point
estimation, this prediction comes along with uncertainty or a marginal
distribution. 

\subsection{Bayesian Optimization}

Bayesian Optimization (BO) offers a framework for finding the extrema of an
objective function:
\begin{equation}
\min_x f(x)
\end{equation}
where $ f(x) $ is reckoned prohibitively expensive to be executed for many
times. Moreover, in the case of black-box problems, one does not have a closed
form but may only query its values at sampled points. As those evaluations
are costly, $ f(\cdot) $ should be invoked as few as possible. 

Any Bayesian method depends on a prior distribution. By thinking of GP as
the prior probability on functions, as analogous to random variables, this prior
represents our beliefs over the space of all feasible objective functions
\cite{Brochu2010}. By doing this, the undetermined function values on non-yet evaluated
points are expressed through a probabilistic distribution. The benefits of this
Bayesian reformulation are obvious. The GP can iteratively incorporate new
information and updates its beliefs on the unevaluated points. These steps
are comparatively cheap. Hence, at each iteration, there is no need to execute
the objective function on all candidate points. We only pick out the most
promising one for the next evaluation according to some criterion. In this way, we restrict
the computations of costly objective function to only on the selected points.
Next, GP updates the posterior beliefs on all the points that are waiting for
evaluations.

A BO algorithm starts with two randomly chosen points. At each step, an
\emph{acquisition function} is maximized to determine the next point to
evaluate, which models the utilities and rewarding of not-yet evaluated points
based on the history information. The next point is then chosen on the $ argmax
$ of this function. The new pair, denoted by $ (x^+, f(x^+)) $, are brought into
a library of evaluated set and reduces the uncertainty around $ x^+ $. The
posterior distribution over the function space is updated and the whole process
is repeated. A nice property is that a BO algorithm can always return a
recommended point if it were to stop without notice.

\subsection{Expected Improvement}

The above procedure of BO holds with a key step, i.e. the selection of a most
promising point. The policy of actively selecting next point highly affects the
performance of an optimizer. A natural and popular one is the \emph{Expected
	Improvement} (EI) \cite{Mockus1975} which estimates the gains of query points
and always selects the one that yields most decrease. In other words, the
selected point is thought to gain the most advance towards the optimum.

To begin with, we first define some notations that would be useful throughout
our paper. Let $ \tilde{x} $ be the best point achieved so far.
\begin{equation}
\label{eq:return:fun}
\tilde{x} =  \mathrm{argmin}_{\mathcal{T} } f(x)
\end{equation}
where $ \mathcal{T} $ denotes the evaluated point set. We represent all points
by $ \Omega $. In principle, the new point to be evaluated, denoted by $ x^+ $, is expected to
lie around $\mathrm{argmin}_{x \in \Omega/\mathcal{T}} {f}(x)$, with $
\Omega/\mathcal{T} $ represents the not-yet evaluated candidate set.

The EI is defined through a relative decrease of $ {f}(x) $ on 
a point $ x $ against the minimal point $ \tilde{x} $ tested till far.
Due to the GP hypothesis on the unexplored points, this quantity
is itself a random variable and is defined as:
\begin{equation}
{I}(x) = \max \{0, f(\tilde{x}) - {f}(x)\}
\end{equation} 
and EI quantity is obtained by taking expectation on both sides. 
\begin{equation}
\label{eq:acq:imp}
\E [{I}(x)] = \E[I(x)|\mathcal{T},x]
\end{equation}

An easy way to compute the above EI acquisition function was derived in \cite{Jones1998}:
\begin{equation}
\label{eq:acq:fun}
EI(x) = {\Sigma} {f(x)} \big(Z \Phi(Z) + \phi(Z)\big)
\end{equation}
in which an intermediate variable $ Z $ is given as $ Z = \frac{{\mu}(x) - f(\tilde{x}) }
{\Sigma(x)} $. $ \Phi $ and $ \phi $ are the standard normal cumulative
distribution function (CDF) and probability density function (PDF),
respectively.

Under only a weak regularity condition, BO based on EI could converge to the
global optimum. Its drawback on the behavioral aspect is that it concentrates
more on exploitation. Points that have high probabilities of being greater than
$ f(\tilde{x}) $ are more preferable than points that with great uncertainty,
even though the latter may have a chance to get larger gains. In other words, EI
is a greedy sampler in a long run. Likewise, Probability of Improvement (PoI),
and Upper Confidence Bound (UCB) are also focusing on maximizing the probability
of rewarding in picking the next point but UCB leaves the trade-off between
exploration and exploitation to the user. The entropy-based policies e.g.
Entropy Search (ES) and Predictive Entropy Search (PES) mainly explore the
unknown areas in order to reduce the uncertainty about the location of the
optimum. In this study, we adopt EI as basic acquisition function. In our
scenario where number of execution iterations is fixed in advance yet usually
insufficient for most BO, the greedy behavior of EI is of least concern to us.

Beside EI, many other acquisition functions have been proposed for various
scenarios. To decide the next point under inequality constraints, Gardner et al.
\cite{Gardner2014} proposed an acquisition function as the product of EI and the
probability of meeting feasibilities of all the constrains. This latter was
realized by assuming the independences among constraints. Other attempts include
modifying the objective function in order to add a penalty on constraint
violations, punishing the unfeasible recommendations made by the acquisition
function. By doing so, these works managed to remove the constraints from the
original problems. For instance, Lee et al. \cite{Picheny2016} constructed an
auxiliary function by making use of augmented Lagrangian formulation. This
technique was extended on handling both the inequality and equality constraints
\cite{Picheny2016}. Bernardo et al. \cite{Bernardo2011} studied the scenario
where the constraint functions are not easily formulated. They worked around
this problem by approximating the unknown function so as to estimate the
probabilities of meeting the constraints. A key update is the proposal of a new
integrated improvement-based criterion conditional on all responses from the
input. This quantity is computed over the entire point space. Likewise, the
acquisition function in \cite{Picheny2014} quantifies the expected reduction of
the uncertainty below the best one found so far. Both quantities need
integrating throughout the feasible points.  Lam et al. \cite{Lam2017,Lam2016}
model the future utilities by looking ahead. The long-term rewards in the
finite-budget scenario are built through dynamical programming over future
steps. However, the lookahead strategy needs more invocations of objective
function otherwise the rewards would come with great uncertainties.

In summary, all the methods are mainly focusing on carefully processing the
objective function and constraints such that they can be both handled in a same
scheme. Few considers the scenario with a finite budget on the number of
invocations of the objective function. In this case, an optimizer needs to be aware
of the remaining budget so as to balance the trade-off between exploration and
exploitation in a principled way.

\section{Directional Constraint}
\label{sec:method}

We are seeking for a BO method, being executed in cases with a limited
evaluation budget, to return as minimal a value as possible. An algorithm is
required to choose samples wisely by assessing the remaining budget. A general
idea is that an algorithm should be more aggressive on exploitation when it is
approaching the evaluation limit. Most probability should be assigned to local
exploitation in the most promising area. This is reasonable as it has undergone
a wide exploration and has a good chance to locate a promising region where an
optimum is likely to be found. However, this intuition cannot be captured by the
simple EI since it will not take the evaluation budget into account.

Motivated by the above analysis, we reformulate the acquisition function with a finite evaluation budget
as a product of EI and a local searching policy (obtained through
considering the neighboring evaluated points). These two terms are balanced
conditional on the remaining evaluation budget. The final acquisition function is
formulated as:
\begin{equation}
\label{eq:trade:off}
H(x)^{\rho(t)} \big[\max\{0, f(x^+) - f(\tilde{x})\}\big]^{1-\rho(t)} \equiv H(x)^\rho EI(x)^{1-\rho(t)}
\end{equation}

$ H(x) $ models the utilities of all possible searching directions pointing from
present $ x_t $, which is denoted as $ g = \frac{x- x_t}{||x - x_t||} $. This
quantity to the power of $ \rho $ is multiplicatively combined with EI. The
power $ \rho $ increasingly changes in the range of $ [0,1] $. An implicit
premise associated with the above formulation is the separability between the
objective function and the directional constraint, which is reasonable since
directions have an analogous role to the gradient. This statistical quantity $
H(x) $ serves a similar role in constraining searching directions in order to
refine the exploring behavior. We refer to this newly proposed directional
statistics as \emph{directional constraint} in order to distinguish it from the
normal ones. Before introducing the design of $ H(x) $, we first look at an
extension and properties of the step function $ \rho(t) $.

As we can observe, EI is a probability quantity measuring the utility of a
sample from the not-yet evaluated point set. Any other probabilistic models are
equally feasible. This makes the constraint widely applicable for a range of
utility models. For instance, in the constrained BO, the utility model could
employ an augmented EI quantity:
\begin{equation}
	EI(x) \equiv C(x) \max\{0, f(x^+) - f(\tilde{x})\} 
\end{equation}
where, the term $ C(x) $ is the product of probabilities of meeting
the feasibilities of equality or inequality constraints.

The trade-off between EI and directional constraint is controlled by the step function $ \rho(t) $,
whose changing pattern affects the behavioral aspects of an optimizer. As we can expect,
the setting of $ \rho(t) $ is problem-dependent and, to some extent, related to
the evaluation budget. In our study, for simplicity, we restrict our discussion
to the simplest design, i.e. $ \rho $ changes linearly with iterations: in $ t
$-th iteration
\begin{equation}
\label{eq:set:fun}
	\rho(t) = \frac{t}{T},\qquad t = \{0,\cdots, T\}
\end{equation}
in which $ {T} $ denotes the number of evaluations. 


\subsection{Modeling Searching Directions}
\label{sec:correction}

We have rectified the EI by augmenting a directional constraint with $ \rho(t) $ as
its exponent. Our remaining issue is to model the utilities of all searching
directions. We again turn to approximation and Bayesian formalism for
a feasible and easy way. However, searching directions are a new type of input
for the BO. Modeling the directional probability as Gaussian distribution
becomes problematic.

Specifically, we are concerned with modeling the adherence to a past direction
or originations in the BO, hoping to achieve a consistent decrease on a local
scale. To do so, we needs to solve two issues: (1) The way to specify an
adherence or consistence to the past direction. (2) A full Bayesian treatment of
directions which relies on devising an iterative update Bayesian rules for
directional statistics.

%

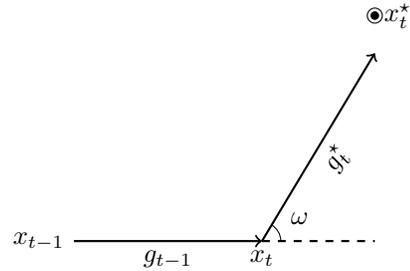
\begin{figure}
	\centering
	\begin{tikzpicture}
	\draw [color=black,thick] (2,3) node[right]{$x^\star_{t} $};
	\filldraw (2,3) circle(0.5mm);
	\draw  (2,3)circle(1mm);
	\draw [color=black,thick,->] (-2,0) node[left]{$x_{t-1}$} --node[below]{$ g_{t-1} $} (0.5,0) node[below]{$x_{t}$};
	\draw [color=black,thick, ->] (0.5,0) -- node[below,sloped]{$ g^\star_{t} $}(2,2.5);
	\draw [color=black,thick,dashed] (0.5,0) --  (2,0);
	\draw (0.75,0) arc (0:60:0.25);
	\node[] at (1,0.3) {$\omega$};
	\end{tikzpicture}
	\caption{This figure illustrates the directional adherence between two
	successful steps in the BO. Provided $  x^\star $ is the ideal point that EI
	suggests and that $ g_{t-1} $ is the last searching direction, a conventional
	searching policy of BO would pick $ x^\star $ since it has the highest
	probability to decrease. However, the uncertainty of GP provides the source for
	jiggles on a local scale. Therefore, in our approach, we complement the
	conventional searching policy by proposing a constraint on the direction of
	actual move. This statistical quantity $ g_{t} $ is demanded to be positively
	correlated with the past origination $ g_{t-1} $. Intuitively, it requires that
	an angle $ \omega $ between $ g_{t} $ and $ g_{t-1} $ is strictly less than 90 degrees.
	\label{fig:adh}}
\end{figure}

Without loss of generality, we assume that two neighboring samples are $ x_t $
and $ x_{t-1} $. The ``optimal'' point recommended by the GP is denoted as
$ x^\star_t$. All points come from a real-valued vector space $ \mathbb{R}^d $.
The direction pointing from  $ x_{t-1} $ to $ x_{t} $ is denoted as $ g_{t} $.
The direction from $ x_t $ to $ x^\star $ is referred to as $ g^\star_t $. The
positional relations between points are illustrated in \cref{fig:adh}.

To begin with, let us revisit a limitation possessed by a conventional searching
policy of BO. When it suggests a point $ x^\star $ based on the estimated
probability from GP and/or the benefit in a long run, due to the smoothness of
GP, we actually get a distribution of ``best'' points. However, the points in
the vicinity of $ x^\star $ are totally neglected by the conventional optimizers, which have
also a great chance get a high reward. This discovery permits us to design a more
effective searching policy by considering the distribution rather than the point
itself.

Due to the uncertainties on the responses, the probability estimated from GP is
severally discounted. This gives the source to the jiggles and
inefficient probing on a local scale. To alleviate this problem, we introduce an additional
constraint for $ g_{t} $, i.e. the angle between $ g_{t} $ and $ g_{t-1} $ is
enforced to be acute. This condition enforces the searching direction $ g_t $
to be adherent to the last one and not reversed, with the hope for 
an adherent optimization path and a resultant consistent decrease. This requirement is illustrated in \cref{fig:adh}. 

To devise an iterative update Bayesian rules, however, is not an easy task as we
are modeling points with GP in an Euclidean space. It is challenging to build a
model on the directions directly. The directional data have a unit norm and are
assumed to live in the surface of a hypersphere. To model this new type of data,
in this study, we make use of Von Mises-Fisher distribution (Vmf)
from directional statistics \cite{Jammalamadaka2001}, which is defined on  $ ( d
- 1 ) $-dimensional hypersphere in $ \mathbb{R}^d $.
\begin{equation}
\label{eq:vof}
Vmf(g|\theta,\kappa) = C_d(\kappa) \exp(\kappa \mathbf{\theta}^T {g})
\end{equation}
where $ ||g|| = 1, g \in \mathbb{R}^{d} $, $ \kappa >0, \kappa \in \mathbb{R} $
and $ ||\mathbf{\theta}|| = 1, \theta \in \mathbb{R}^d $. The parameters $
\theta $ and $ \kappa $ represent the averaged direction and concentration
parameter, respectively. In particular, the greater the value of $ \kappa $, the
higher concentration the probability mass is around the mean direction. The
distribution is uniform on the sphere when $ \kappa = 0 $. The normalization
constant $ C_d(\kappa) $ satisfies
\[ \int_{||{x}||=1}\exp(\kappa {\theta}^T {x})\,d\mathbf{x} = \frac{(2\pi)^{d/2-1} B_{d/2-1}(\kappa)}{\kappa^{d/2-1}}\equiv C_d(\kappa) \] where $ B_p $ is a modified Bessel function of order $ p $. If $ d = 2 $ the distribution reduces to the von Mises distribution (aka. circular normal distribution) on a circle.

We take termination from Bayesian formalism. The first step is to cast the
points in the Euclidean space to the elements on the sphere. This is easy as we
are dealing with points along an optimization trace. The directional statistics
is built through the directions along the optimization trace as shown in
\cref{fig:adh}. The prior probability on $ g_{t} $ is assumed to be centered
around $ \theta_t $ with concentration parameter $ \kappa_t $:
\begin{align}
 H_t(x) \triangleq Vmf( g_{t} |\theta_t, \kappa_t)
\end{align}
where $ g_t = \frac{x - x_{t-1}}{||x - x_{t-1}||} $. Note that the time
stamp is added as subscript to $ H(x) $ for clarity.

We approximate the posterior distribution $ p(g_{t+1}|x_{t+1}) $ via a function of same form $
Vmf(g_{t+1}|\theta_{t+1},\kappa_{t+1}) $, where $ \theta_{t+1} $ and $
\kappa_{t+1} $ are unknown parameters waiting to be inferred. 

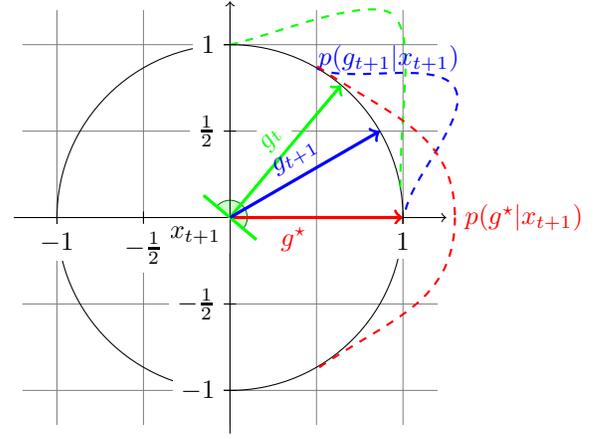
\begin{figure}[t]
	\centering
	\begin{tikzpicture}[scale=2.3]
	\draw[step=.5cm, gray, very thin] (-1.2,-1.2) grid (1.2,1.2); 
	\filldraw[fill=green!20,draw=green!50!black] (0,0)-- (-40:0.1) arc (-40:140:0.1) -- cycle; 
	\draw[->] (-1.25,0) -- (1.25,0) coordinate (x axis);
	\draw[->] (0,-1.25) -- (0,1.25) coordinate (y axis);
	\draw (0,0) circle (1cm);
	\draw[very thick,red,->](0,0)--node[below left,fill=white]{$g^\star$}(0:1);
	\draw[very thick, green,-](140:0.2) -- (-40:0.2);
	\draw[very thick, green, -> ](0,0) -- node[above,sloped,fill=white]{$ g_t $}(50:1);
	\draw[thick, green, dashed](90:1)..controls++(10:0.2) and ++(120:0.2)..(50:1.5);
	\draw[thick,green, dashed](10:1)..controls++(90:0.5)and ++(-60:0.2)..(50:1.5);
	\node[blue] at (45:1.3){$ p(g_{t+1}|{x}_{t+1}) $};
	\node[red] at (0:1.7) {$ p(g^\star|{x}_{t+1}) $};
	\draw [thick,blue,dashed] (60:1)..controls ++(150:-0.2) and ++(120:0.2).. (30:1.5);
	\draw [thick,blue, dashed] (30:1.5)..controls ++(120:-0.2) and ++(90:0.2).. (0:1.01);
	\draw[thick,red,dashed](60:1.01)..controls++(145:-0.5) and ++(90:0.5)..(0:1.3);
	\draw[thick,red,dashed] (0:1.3)..controls++(90:-0.5) and ++(35:0.5)..(-60:1.01);
	\draw[very thick, blue,->] (0,0) -- node[above,sloped]{$ g_{t+1} $}(30:1cm);
	\node[below left, fill=white](0,0){$ {x}_{t+1} $};
	\foreach \x/\xtext in {-1, -0.5/-\frac{1}{2}, 1} 
	\draw (\x cm,1pt) -- (\x cm,-1pt) node[anchor=north,fill=white] {$\xtext$};
	\foreach \y/\ytext in {-1, -0.5/-\frac{1}{2}, 0.5/\frac{1}{2}, 1} 
	\draw (1pt,\y cm) -- (-1pt,\y cm) node[anchor=east,fill=white] {$\ytext$};
	\end{tikzpicture}
	\caption{A schematic diagram to infer the posterior distribution of searching
		directions at step $ t+1 $. In this diagram, the past direction $ g_t $ in green
		and the direction suggested by GP in red are synthesized to compute the
		posterior distribution for $ g_{t+1} $. In order to maintain the adherence and
		consistence to an optimization path, the inconsistent part in the distribution of
		$ g^\star $ is pruned, with the hope to avoid the jiggles in the local search.
		\label{fig:post}}
\end{figure}

Moving from $ x_t $ to $ x_{t+1} $, GP updates our belief about the response
surface in the region surrounding $ x_{t+1} $ and delivers an updated $
x^\star_{t+1} $. We should modify the searching direction $ g_{t+1} $
accordingly. Hence, at time $ t+1 $, we have the past orientation $ g_t $ and
updated $ g^\star_{t+1}$ which starts from $ x_{t+1} $ and ends at $
x^\star_{t+1} $. We turn to Bayesian rules with the hope to synthesize two
knowledges. The posterior distribution is computed as follows:
\begin{align}
\label{eq:posterior}
H_{t+1}(x)& = p(g_{t+1}) \\
&= \int d g_t^\star d g_t \I(\langle g_{t+1}, g_t\rangle \ge 0) p(g_t) p(g_{t+1}|g^\star_t)p(g^\star_t) \notag
\end{align}
where $ g_{t+1} = \frac{x - x_t}{||x - x_t||} $.

In the above \cref{eq:posterior}, we enforce the adherence to hold by adding condition $ \I (\langle
g_t, g_{t+1}\rangle \ge 0) $, where $ \I(\cdot) = 1 $ when $ \langle g_t,
g_{t+1}\rangle \ge 0 $ holds, otherwise $ \I(\cdot) = 0 $. We illustrate
this idea in \cref{fig:adh}. The point $ x^\star $ is the point GP suggests on
step $ t+1 $. These two points $ x_t $ and $
x_{t+1} $, together with $ x^\star_{t+1} $ consist an optimization path. The
jiggle happens when the new searching direction $ g_{t+1} $ deviates much from
the past orientation or even reverses to $ g_{t} $. This condition exists
to alleviate the jiggles in the update rules for $ g_{t+1} $ by only considering
$ g_{t+1} $ that spans an acute angle with $ g_{t} $. Overall, $ H(x) $ guarantees
that these two steps passing through are adhering to an optimization path and thus may
lead to a consistent decrease. This general idea is graphically illustrated in
\cref{fig:post}.


The parameter $ \kappa_{t+1} $ of VMF represents how the directions disperse on the
sphere. As we have already noticed, there is no
closed-form for the posterior estimation of $ \kappa $. We again refer to
approximation for an easy way to compute this random quantity. To facilitate
presentation, we move the technically cumbersome proof to Appendix. An easy way
in the 2D case is presented below.

\begin{figure*}[!t]
	\begin{center}
		\subfloat[\label{fig:pri:p1}
		]{\includegraphics[width=0.35\linewidth]{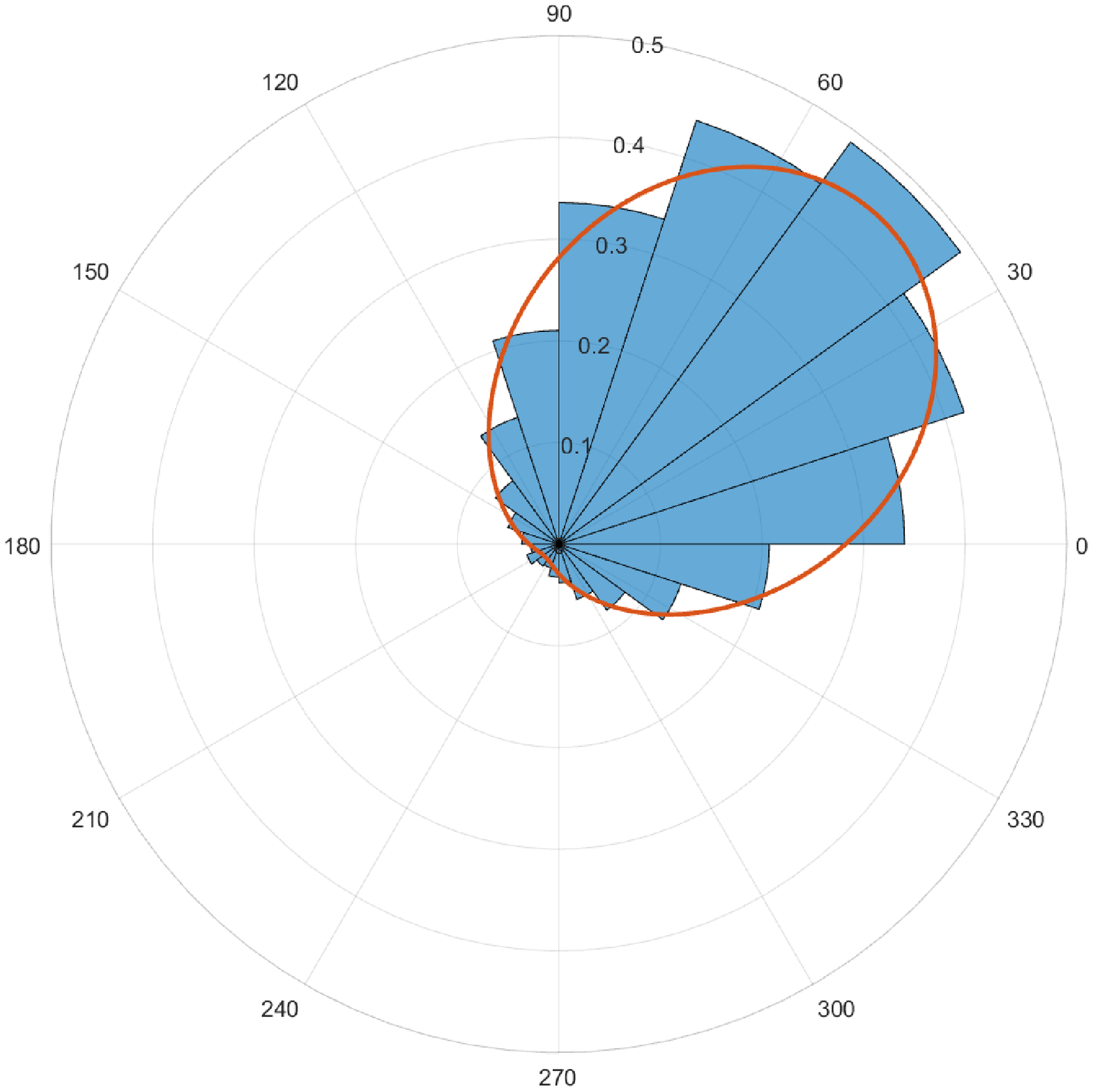}}
		\hfil \subfloat[\label{fig:pos:p2}
		]{\includegraphics[width=0.35\linewidth]{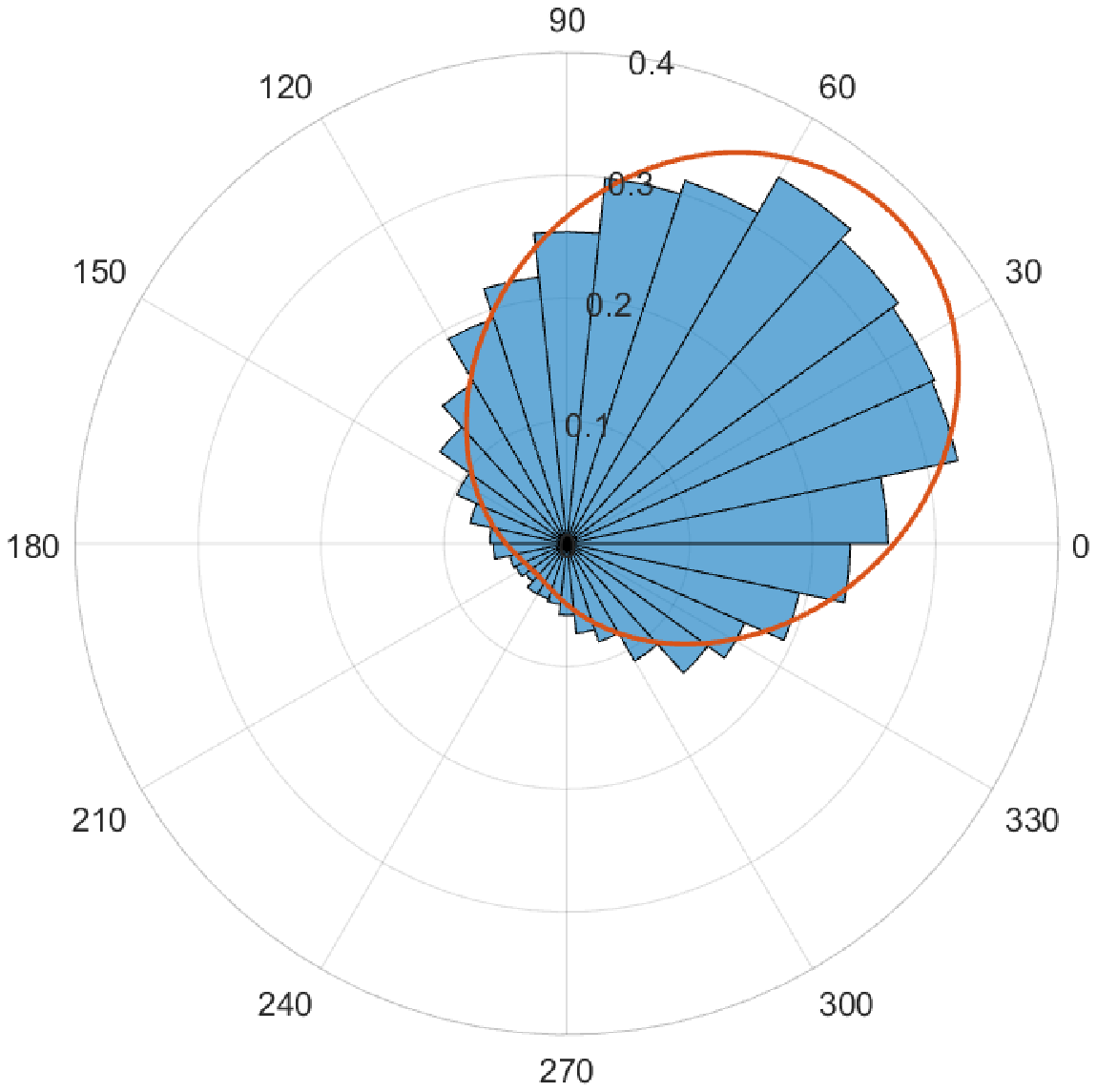}} \caption{ The
			circular histogram for both prior and posterior distribution. The left panel (a)
			displays the statistics along with the true circular distribution enveloping
			the bars. The right panel (b) displays the approximated posterior distribution in the form of $ Vmf $. From the figure, it can be concluded that, in our current
			experimental setting, $ Vmf $ is an appropriate PDF choice. Moreover, the
			approximated probabilities through Bayesian rules matches the true posterior distribution.}
	\end{center}
\end{figure*}

When $ d = 2 $, corresponding to the case we are optimizing in a flat plain. The parameters  $ \theta_{t+1} $ and $ \kappa_{t+1} $ are updated as:
\begin{align}
\label{eq:approx:dim2}
\begin{split}
\kappa_{t+1}& = \frac{1}{4} \sqrt{\hat{\kappa}^\star \kappa_t \left(\hat{\kappa}^\star{}^2+\kappa_t^2 -8 + 16 k_t^2\right)}\\
\theta_{t+1}& = \frac{k}{\sqrt{1+k^2}}g^\star_{t+1} + \frac{1}{\sqrt{1+k^2}} g_t
\end{split}
\end{align}
where $ k =  \frac{1}{4 k_t} \sqrt{\hat{\kappa}^\star \kappa_t
	\left(\hat{\kappa}^\star{}^2+\kappa_t^2 -8\right)}$ balances between $
g^\star_{t+1} $ and past direction $ g_{t} $. More details including the
descriptions of variables are given in Appendix. In principle, these update rules make $ \kappa
$ self-adaptive, which could be quite helpful when the algorithm is exploring a
rugged target surface.

\section{Experiment} 
\label{sec:exp} 

We evaluate our method on several problems:
three synthetic tasks and two real-world datasets. We implemented our method on
the basis of GP. All GP hyper-parameters were selected under the criterion of
maximizing the marginal likelihood.

\subsection{Evaluation Criterion}

To compare the performances of different Bayesian optimizers, we use utility gap metric
\cite{Hernandez-Lobato2015}. At iteration $ n $, this metric measures the gap
between the optimal function value $ \min(f) $ and the value of objective
function at a recommended point $ \tilde{x}_n $ of step $ n $. Recall that $ \tilde{x}_n =
 \mathrm{argmin}_{\mathcal{T} } f(x) $, where $ \mathcal{T} $ is
evaluated point set till iteration $ n $.
\begin{equation}
e_n = 
	\begin{cases}
	|f(\tilde{x}_n) - \min(f) | & \text{if } \tilde{x}_n \text{is feasible.} \\
	| \Psi - \min(f) | & \text{otherwise.}
	\end{cases}
\end{equation}
where $ \Psi $ a penalty function when the optimizer recommends an infeasible
point or makes no recommendation. $ \Psi $ could be any large real-valued constant greater than $ \max f(x) $. The recommended point $ \tilde{x}_n $ is
different from the one selected for testing at step $ n $. The recommended point
is the best point seen so far by the optimizer. In other words, if the optimizer
were stopped at step $ n $, this point would be the most promising one. If
the function optimum $ \min(f) $ is inaccessible, we use the magnitude of
function at $ \tilde{x}_n $ instead as we are doing minimization.

\subsection{Synthetic Problems} 
\label{sec:syn}

\textbf{Simulation 1}

We begin with one synthetic data generated according to VMF distribution. We
verify the agreement between the Bayesian update rules (detailed in Appendix) and the
results estimated from samples. This numerical experiment is to verify
the precision as we make approximations during the calculation of posterior
probability. In this simulation, the prior knowledge of $ g_t $ took the form
of $ Vmf $, whose $ \kappa = 1 $ and mean direction as $ \theta=1 $. The $
g^\star $ was sampled from an isotopic Gaussian distribution with $ [1,1]^T $ as
its mean vector and $ [1.5,1.5]^T $ as the covariance diagonals. We took 10000
samples from this distribution to estimate the intermediate variable $ \hat{R} $. The posterior distribution was assumed to take the form of $
Vmf(\cdot) $. $ g_{t+1} $ is generated under the constraint that $ \langle
g_{t+1}, g_{t} \rangle > 0$, such that we do not need to consider the trivial
case in \cref{eq:posterior}. The circular histogram depicts the distribution of
samples in a polar coordinate system. The enveloping curve around the circular
histogram indicates the continuous probability density function (PDF) computed
through Bayesian approximation. We scaled the PDF for the convenience of comparison.

By looking at the \cref{fig:pos:p2}, one can observe two phenomenons: (1)
Loosely speaking, the observed orientations generated under the constraint of
acute angle normally coincide with the $ Vmf $. Hence, it is reasonable for the
assumption for the PDF of post-data probability being taken as $ Vmf $, at least
in this simplified simulation setting. (2) The PDF calculated by our method, as compared with sample histogram, confirms the general agreement
between the estimated distribution and \emph{a posteriori} circular histogram
for observations, although not ideally. If considering the computational
efficiency, the estimated PDF used in this scenario exhibits a great advantage.
We will return to investigate the effects of approximations on the overall
performance in later simulations.

\textbf{Simulation 2}

In this simulation, we adopt one synthetic problem, which is a two-dimensional
continuous and multi-modal function. It has been commonly used as a benchmark problem
for global optimization methods \cite{Gardner2014}. In this experiment, an optimizer
is searching over the response surface for the global minimum.
Its effectiveness is evaluated not merely through whether this
optimizer manages to locate the optimum within a limited execution budge,
but also depends on how it balances the exploitation and exploration within the
execution process to speed-up the searching. The latter obviously determines its overall performance. 

In this simulation, the execution budge was setup to be fifty iterations, or at most fifty invocations of target functions. The searching is restricted
within a rectangular area, $ [-5,0] \times [-5,5]$. The response surface is a 2D
plate, and the objective function is described by the following formula:

\begin{equation}
\label{eq:syn:obj1}
	\mathcal{O}(x,y) = \cos(2x) \cos(y) + \sin(x)
\end{equation}

Besides, this problem could be more challenging by adding a constraint that
splits the rectangular area into multiple feasible bands. The constraint is
formulated as in \cref{eq:constraint}.
\begin{equation}
\label{eq:constraint}
	c(x,y) = \cos(x) \cos(y) - \sin(x) \sin(y) \leq 0.5
\end{equation}

After imposing the above constraint, the optimum is found to be near the edge of
a feasible band, which makes it a challenging problem as it is no longer
smooth near the optimum. In this case, as an optimizer has little knowledge about the
feasible band, it is inclined to consume execution budget on probing the
feasible boundary whilst searching for the optimum.

\begin{figure}[t]
	\centering
	\includegraphics[width=0.8\linewidth]{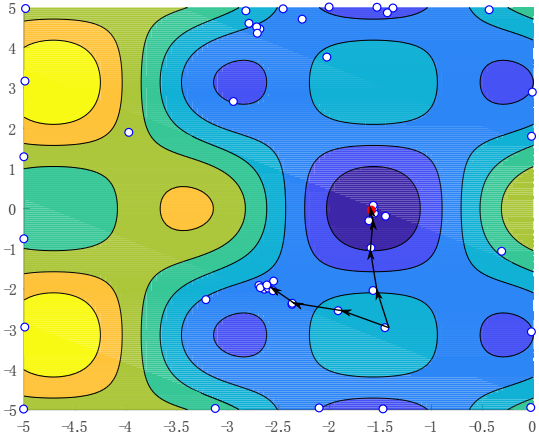} \caption{The DCBO trace
		under no constraint. Fifty evaluations are permitted. We connect two successive
		optimization paths by arrows. We can see that, near the end of
		iterations, the searching does not disperse over the whole space but
		reduce to a local behavior.\label{fig:bo}} 
\end{figure}
In this simulation, we first investigate the behavior of DCBO without the
additional constraint \cref{eq:constraint}. In this setting, an algorithm is facing a squared feasible region, and it needs to make a recommendation within fifty
trials, which is normally tight for an
optimization problem. An execution trace is depicted in \cref{fig:bo}. It is
interesting to note two properties on its behavior: (1) The algorithm began
with a random search for the global optimum by aimless probing on a 2D response
surface. This is common in all Bayesian optimizations due to current lack of
knowledge. (2) However, as the algorithm run, it learned about locations of 
probable optimum with a strong confidence. It fast concentrated its searching into the most possible areas. (3) In this example, besides
the centric area it explored, it also searched the upper part since this 
also appears to be promising judged from history information. Its behavior
exhibited jiggles since it had enough budget and directional constraint did not take great
efforts in the beginning. DCBO erroneously converged to a local optimal point but managed to
escape from it. 
\begin{figure}[t]
	\centering \includegraphics[width=0.8\linewidth]{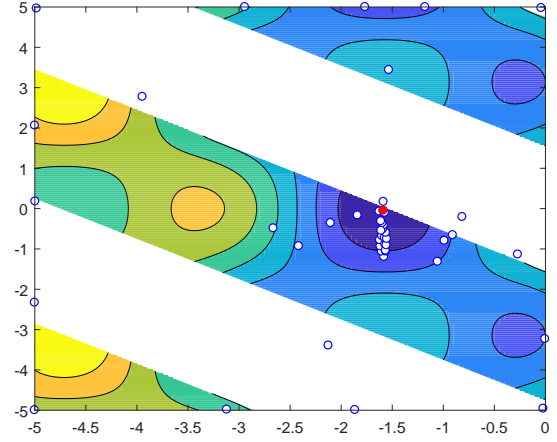}
\caption{The DCBO traces under a constraint. The blank areas indicate the
	infeasible regions. As before, fifty evaluations in total were allowed. We can observe
	that the directional constraint has a benefit to make new samples be prone to
	stay in the feasible region without asking for an explicit condition of
	feasibility. 	\label{fig:cbo} }
\end{figure}

\begin{figure*}[th]
	\centering
	\centering
	\subfloat[\label{fig:cbo:cangle:media}]{\includegraphics[width=0.4\linewidth]{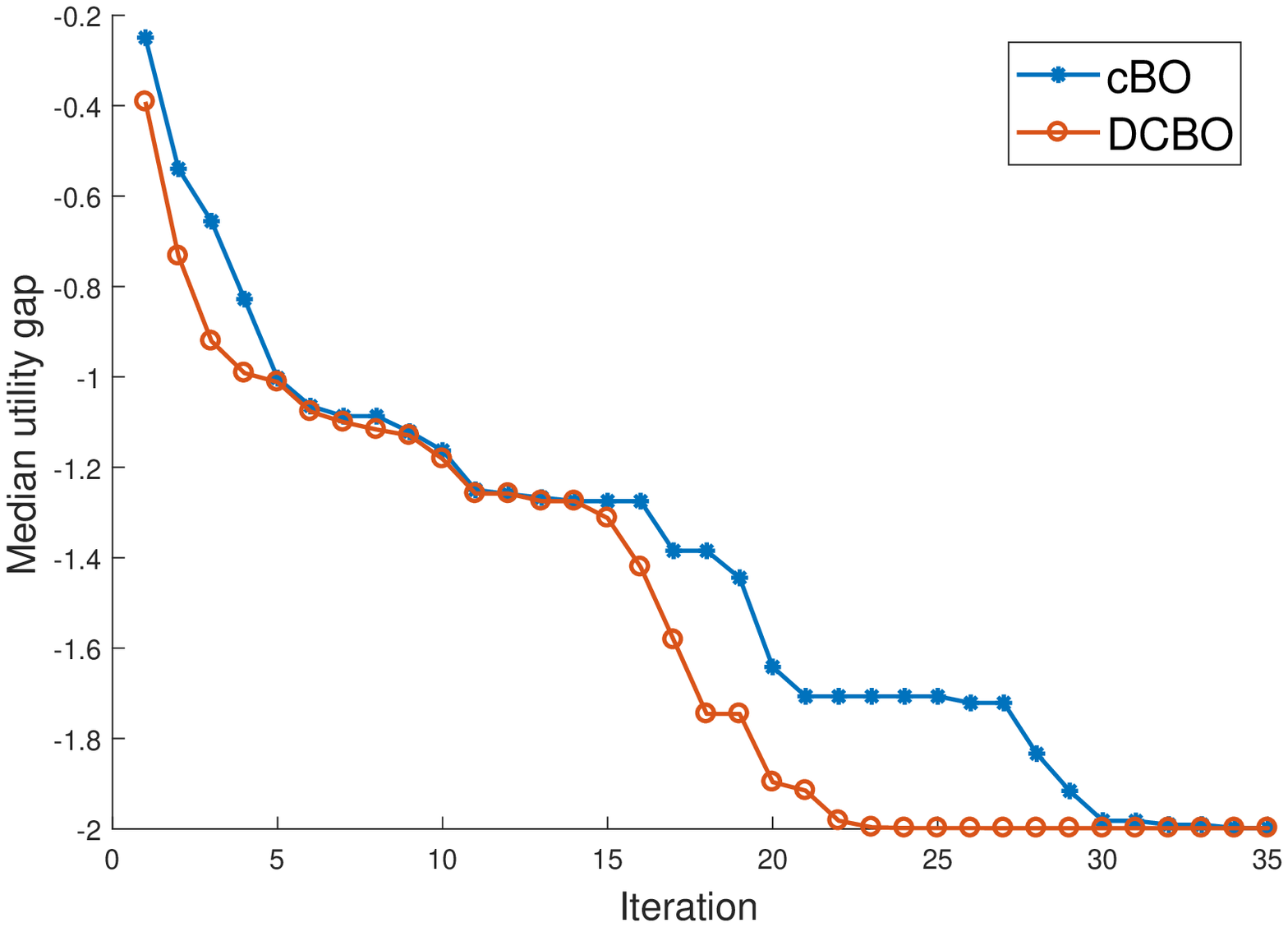}}\hfil
	\subfloat[ 	
	\label{fig:cbo:cangle:mean}]{\includegraphics[width=0.4 \linewidth]{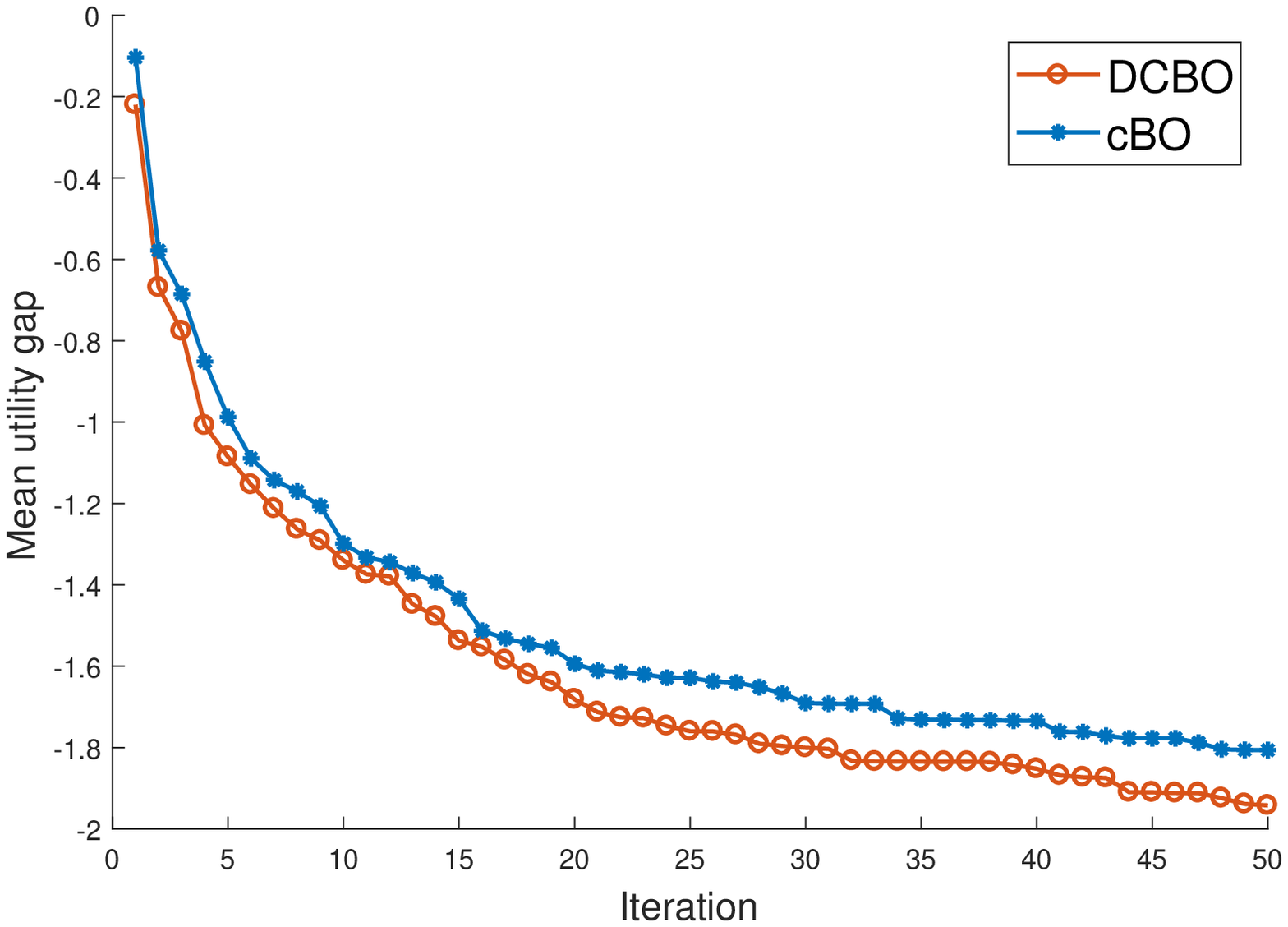}}
	\caption{(a): The comparison between cBO and DCBO in terms of median utility gap
		metric over 50 runs. For clarity, we only
		display the results up to 35 iterations. The curves indicate that both
		optimization policies behave similarly from the beginning. However, as the DCBO
		integrates additional directional knowledge inferred from its past samples, it
		gains a relative advantage, thus leading to a better overall performance. (b): The comparison between cBO and DCBO in terms of mean utility gap metric. The result shows that the directional constraint tends to increase the probability for an algorithm to return a better recommendation under a given evaluation budget. \label{fig:cbo:cangle}}
\end{figure*}

\begin{figure*}[!ht]
	\centering
	\subfloat[\label{fig:bo:poi:median}]{\includegraphics[width=0.4\linewidth]{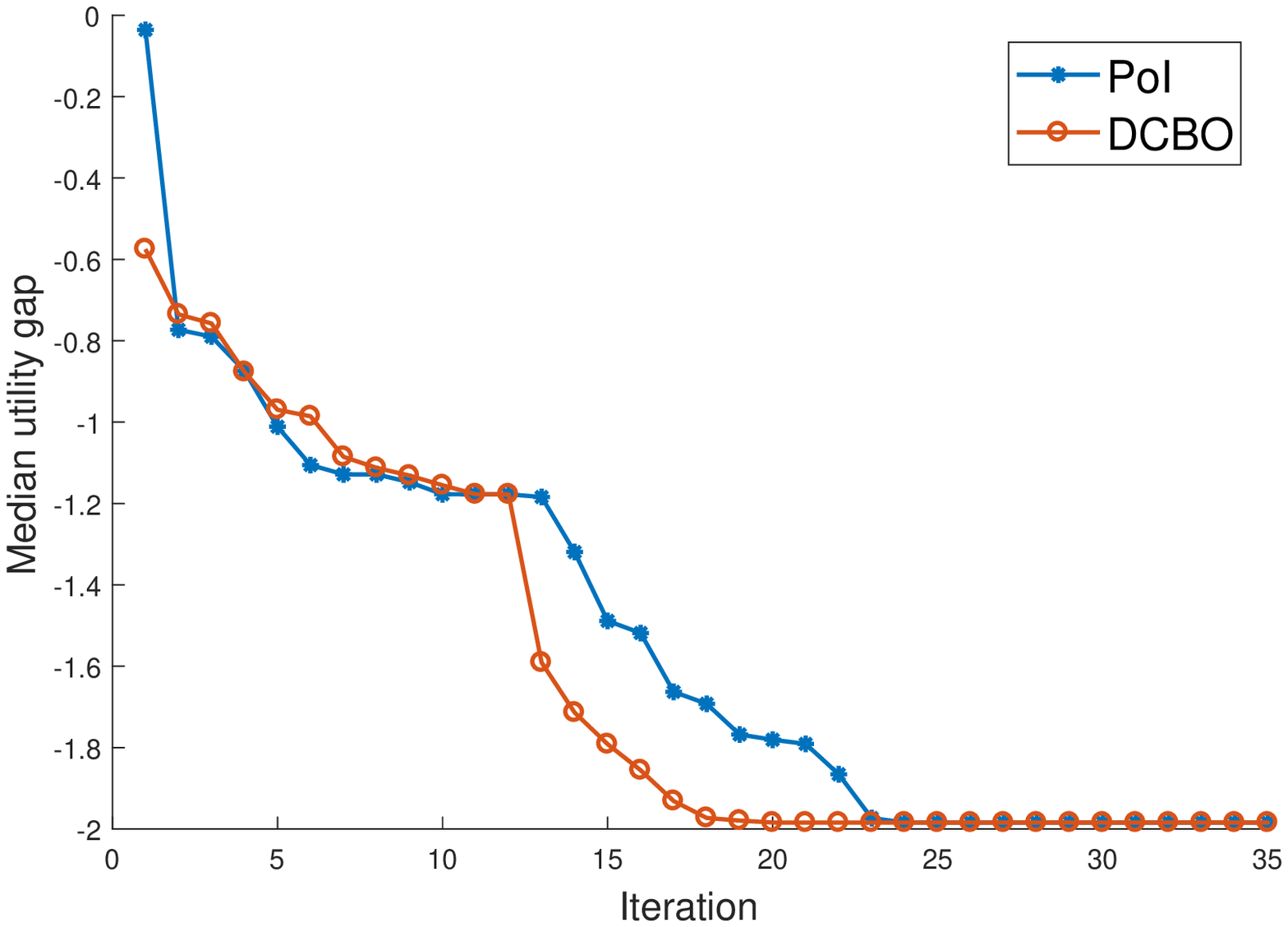}} \hfil
	\subfloat[\label{fig:bo:poi:mean}]{\includegraphics[width=0.4\linewidth]{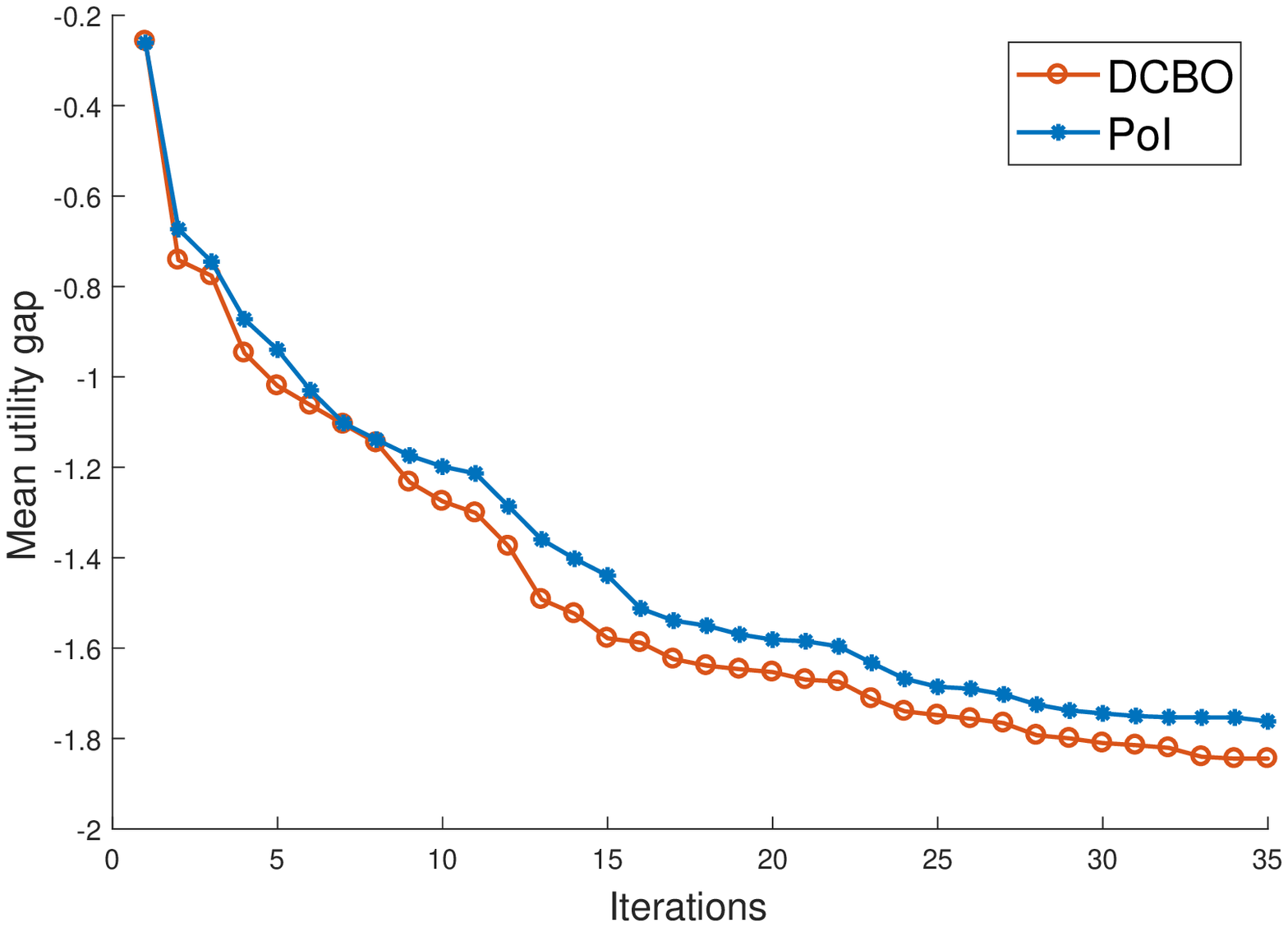}}
	\caption{(a): The comparison between PoI and DCBO in terms of median utility gap measures
		over 50 runs. The curves demonstrate that, after an
		initial search, DCBO can arrive at the global optimal point more quickly than PoI. (b): The comparison between PoI and DCBO in terms of mean utility gap measures. \label{fig:bo:poi}}
\end{figure*}

\begin{figure*}[ht]
	\centering
	\subfloat[\label{fig:ei:data:1}]{\includegraphics[width=0.4\linewidth]{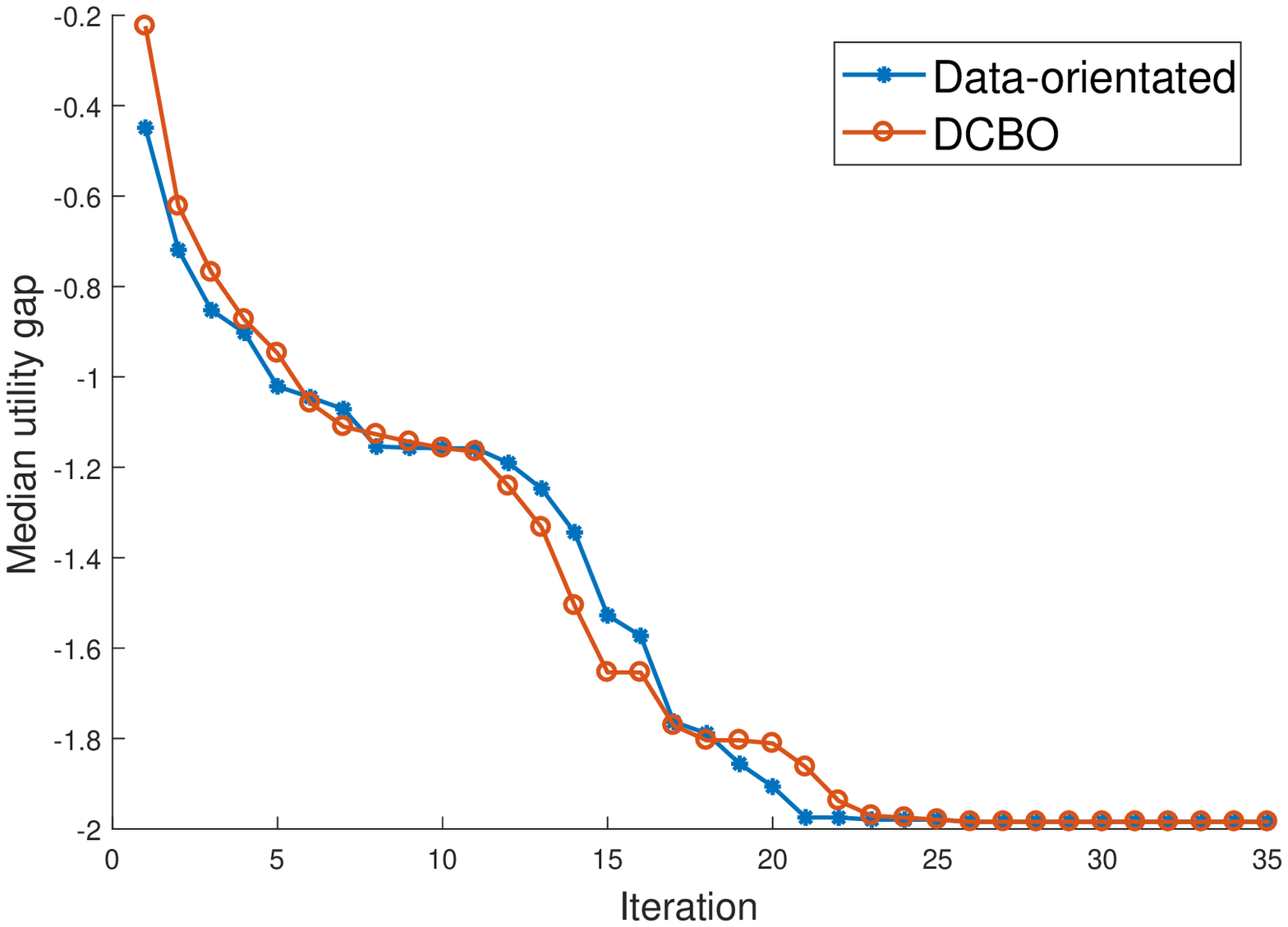}} \hfil
	\subfloat[\label{fig:ei:data:2}]{\includegraphics[width=0.4\linewidth]{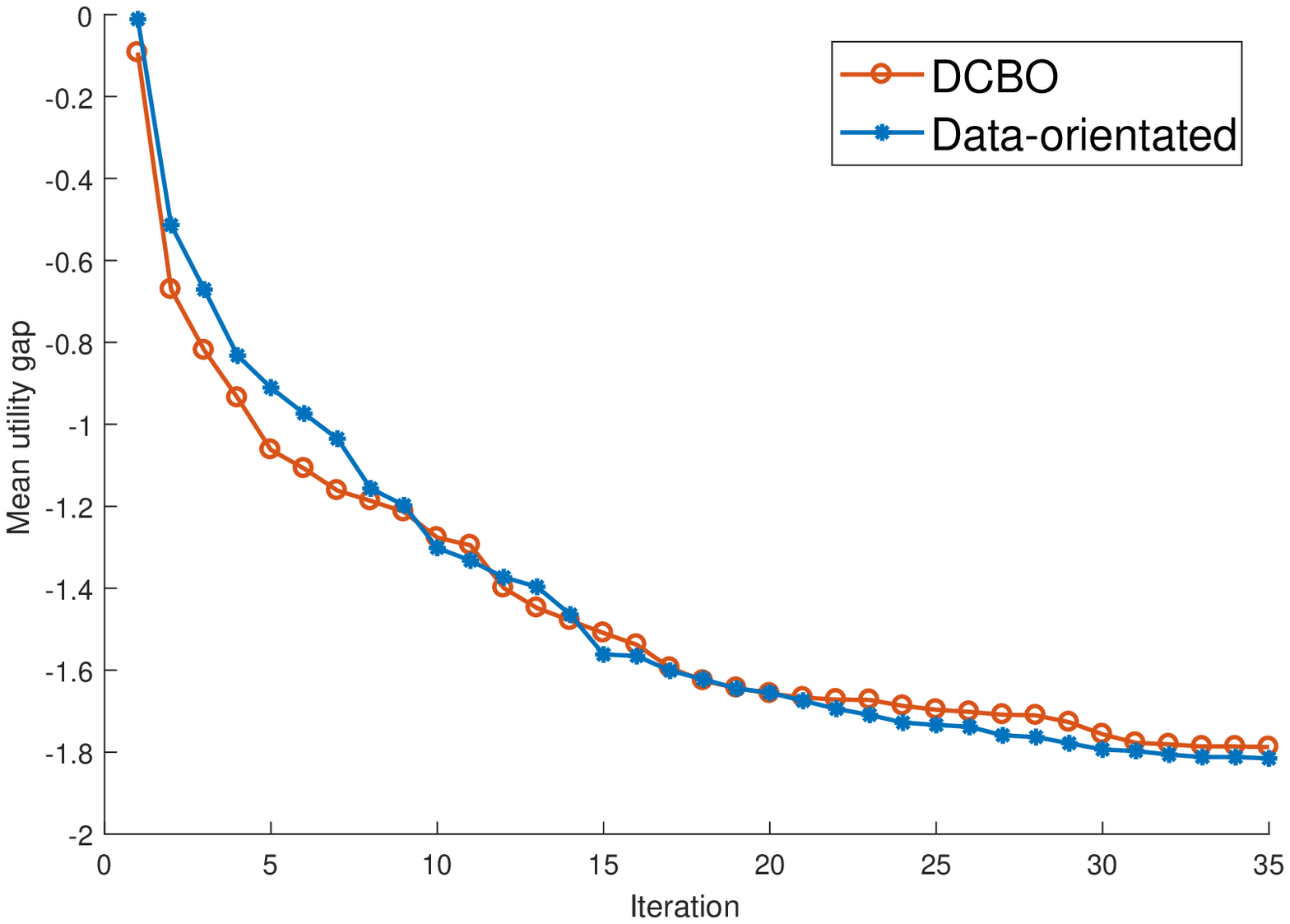}}
	\caption{ (a): The comparison between the data-orientated approach and the DCBO in terms of
		median values of utility gaps. (b): The comparison between the data-originated  approach and the DCBO in terms of mean values of utility gaps.
		From this comparison, we know that two approaches have quite a close behavior in most cases.  More details on set-up can be found in the text.\label{fig:ei:data}}
\end{figure*}

In \cref{fig:cbo}, we elucidate an execution trace on the constrained problem.
The effects of directional constraint on its behavioral aspects are more evident. The experimental setting keeps identical
to the above example. From the figure, we can observe that: (1) This optimizer
did not consider the upper part of figure as a promising area, so, in contrast
to the above example, it did not take efforts to search for an optimum there.
This phenomenon is not typical since each algorithm was initialized randomly. It devoted much
of its model capacity to the centric region, which, as it appears, is the most
promising area. The evaluation limit was relatively adequate when an optimizer
happened to identify the best searching region. The redundant trials lead to
more local searches. (2) A less noticeable phenomenon is that although it has an
adequate budget, an optimizer has very few infeasible samples off the feasible
boundaries. This is due to the correction on searching directions when an
infeasible samples were made. This correction guarantees that the next sample is from a more
judicious searching direction.

We will demonstrate in \cref{fig:cbo:cangle} that the above analyses are not special. As
suggested in \cite{Hernandez-Lobato2015}, median value would be a reasonable representative
for the mass of the data. Since each algorithm was initialized independently and
randomly, the execution traces exhibit disparities. Thus, the empirical
distribution of the utility gap measurements is heavy-tailed. An appropriate
representative in this case would be their median values. But for completeness, we also report the
averaged values as in \cref{fig:cbo:cangle:mean}.

In this simulation, we first compare the performances between constrained
Bayesian optimization (cBO) \cite{Gardner2014} and DCBO proposed in our study.
The former approach, cBO, is specially proposed for such an optimization problem
under inequality constraints and has been served as a benchmark.

By looking into \cref{fig:cbo:cangle:media}, the curve exhibits a typical behavior of
DCBO. As expected, the first half part (before iteration 15), both two curves have a highly similar behavior. In this stage,  both are concentrating on 
exploring the unknown regions. After uncertainty over the response surface has
been reduced significantly, DCBO decreases more quickly than its counterpart,
achieving an overall advantage within the evaluation budget. Its counterpart,
cBO, experiences two more plateaus before it reaches at optimal point. The
plateaus, if shown in a figure, represents the inefficient jiggles on the
response surface. A
similar result in \cref{fig:cbo:cangle:mean} demonstrates that DCBO achieves a consistent advantage over cBO as the two
algorithms run.

In addition to the comparison with cBO, we also conduct the experiment between
Probability of Improvement (PoI) and DCBO on the unconstrained problem. PoI is a
sampling policy that selects the next sample according to the probability of
possible improvement brought by a candidate point. We evaluated PoI and DCBO on
the problem as described by \cref{eq:syn:obj1} but without constraint as 
\cref{eq:constraint}. The total evaluation budget is still fifty times.

The experimental results are shown in \cref{fig:bo:poi}. In \cref{fig:bo:poi:median}, a considerable speed-up is
achieved as compared with the PoI. The behavioral aspects are
analogous to that in \cref{fig:cbo:cangle:media}. The mean curves in
\cref{fig:bo:poi:mean} report a similar result. In current experimental
settings with a limited evaluated budget, DCBO is superior to PoI.

\textbf{Simulation 3}

In this simulation, we examine the approximated solution provided as in
\cref{eq:approx:dim2} and the one estimated from sampled data. We check whether the
technique of approximation may result in a performance degradation and thus slow down the optimization progress. Rather than
assuming the form of probability distribution of directional data and making
an approximation such that the posterior probability still keeps tractability, we drop these assumptions and directly estimate the necessary
quantities from observations.

In this simulation, at time $ t $, we sample a ``best'' point $ x^\star $
according to its posterior distribution inferred from GP. Every $
x^\star $ will result in a possible direction starting from current point $ x_t
$. If it spans a acute angle with the previous searching direction, this one is
accepted, otherwise dropped. This procedure is repeated for 1000 times. The directional
estimations are numerical normalized and interpolated over all directions. As
indicated above, this method is numerical expensive and has a low efficiency.

The approaches to be compared are: (1) the approximation approach that is
developed based on Bayesian formalism, that is, the DCBO; (2) the approach that
have its posterior distribution numerically estimated via the sampling, which we
call data-oriented approach. The experimental results of two approaches are
displayed in \cref{fig:ei:data}. Their
performances are measured through the median and the mean values of utility
gaps. By checking the two figures, we can see that, although data-orientated
approach does not have hypotheses on the form of posterior probability and
interpolates probabilistic quantities using expensive numerical samplings, there
is no significant advantage over DCBO, which is developed based on
the approximation. Moreover, as shown in \cref{fig:ei:data:1}, two curves exhibit
a nearly identical tendency and are quite close to each other, representing
that, in most cases, DCBO behaves almost the same as the data-orientated
approach. A same analysis is also applicable to \cref{fig:ei:data:2}.

\subsection{Benchmark task}

We assess DCBO by learning latent (hidden) parameters for linear dynamical
system (LDS). Because of its mathematical analyzable structure, and predicative
behavior, a time-invariant discrete LDS is arguably the most commonly used model
for real-world engineering applications. Many physical systems could be
accurately described by this model. The exact inference within the LDS can be
done via Expectation Maximum based (EM-based) approaches. EM is an iterative
optimization algorithm which requires many iterations to converge. During each
iteration, an algorithm needs to pass over the entire dataset. Hence, EM-based
algorithms do not scale well to very long observations. Moreover, the parameter
optimization in EM use either gradient descent or its variants and also needs many
iterations to converge. These two factors together make EM-based
approaches impractical for large dataset.

In this real-world application, we adopt an EM-based approximation method called
Approximate Second-Order Statistics (ASOS) for learning the parameters of an
LDS. This method was designed to trade-off a high-quantified solution
against a speed-up on execution, which is controlled by a meta-parameter $ klim $.
This parameter has a broad feasible region. We restrict our attention to $
klim $ and the latent space dimension, as both are critical in determining its 
model capability. Carefully tuning these parameters can be prohibitively
expensive. In contrast to the previous simulations, this
problem is a discrete optimization problem.

In this experiment, EM-based approach was tested on a parametric combination and returns its 
negative log likelihood. This quantity measures the utility of this combination. As
EM-based algorithm is low efficient, in our experimental setting, the maximum permitted number of executions is 50 times. An optimizer that quickly converged to a
low negative log likelihood was valued high. All runs were initialized
with different initial parameters, which were generated randomly and projected
into their feasible regions. The optimization were repeated for fifty times. The
mean values of utility gaps, as well as standard deviations, were collected and reported.
As described in \cite{Martens2010}, every 10 iterations, an exact log likelihood
was computed to revise the approximated ones. 

We compared DCBO against several searching polices: PoI, EI, UCB with various $
\kappa $. Some of them are among widely-used policies in
real-world applications, due to their simplicity on conception and predictive
behaviors. In particular, UCB
leaves this task of trade-off to practitioners. Its parameter $ \kappa $ changes
the proportion of exploration in one search. We vary this parameter in a
representative set $ \{0.5, 1, 6, 9\} $. Larger value means more exploration for
UCB.


We adopt two real-world datasets. The first we used in our experiment was a
3-dimensional sequential data of length 6305. This dataset\footnote{This dataset
	is a standardly used for identifying a system and can be accessed via
	\url{http://homes.esat.kuleuven.be/~smc/daisy/daisydata.html}.} was collected by
sensors from an industrial milk evaporator. The second dataset is transformed
sensor readings of length 15300 on motion capture. We used the first ten
dimensions of this 49-dimensional dataset \footnote{This dataset is available
	on-line from \url{http://mocap.cs.cmu.edu/}. The preprocessing step was
	described as in \cite{Taylor2007}}. Two datasets are referred to as \emph{milk} and \emph{motion}, respectively. 

\begin{figure}
	\centering \includegraphics[width=.9\linewidth]{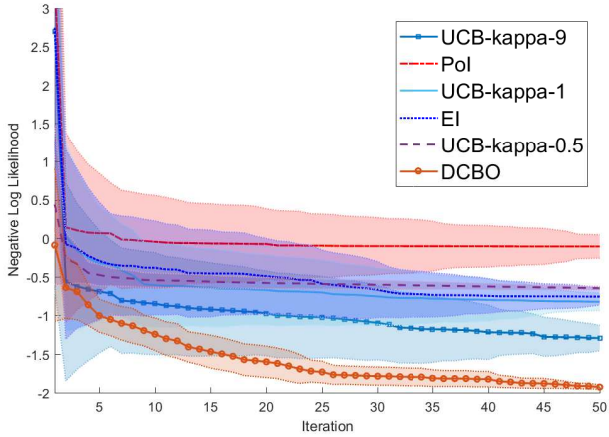}
\caption{ The negative log likelihood returned by Bayesian searching policies.
	They are executed on the task of inferring the parameters for an LDS on the first
	dataset, \emph{milk}. The mean and standard deviation are reported for each
	searching policy. On this challenging discrete optimization problem, DCBO reports a
	superior performance in recommending better parametric combinations for the
	LDS. From the figure, many searching policies are hindered by flat optimization
	plateaus and slow down. DCBO, however, keeps a steady
	decreasing tendency even in comparison with UCB-kappa-9 which spends more on exploration.
	\label{fig:benchmark:1} }
\end{figure}

\begin{figure}
	\centering \includegraphics[width=.9\linewidth]{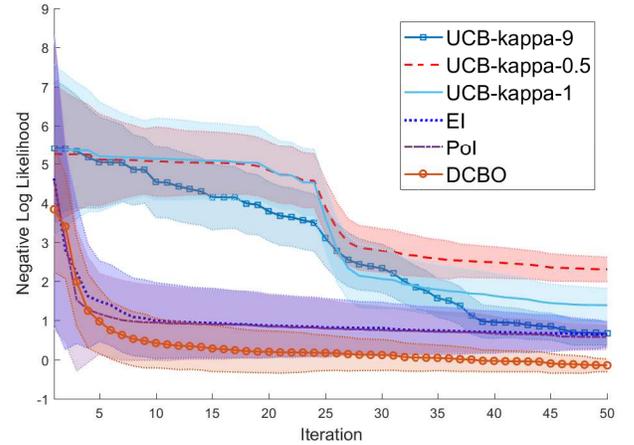}
	\caption{ The negative log likelihood returned by Bayesian searching policies.
		They are executed on the task of inferring the parameters for an LDS on the second
		dataset, \emph{motion}. On this dataset, DCBO is the most effective one. Although
		UCB-kappa-9 finally achieves a comparable result, it has a large deviation
		and experiences a long stagnation where no significant progress is made. EI and
		PoI have analogous traces and nearly overlap each other. Not surprisingly, DCBO
		preserves their major features on shapes but accelerates the decreasing pace.
		\label{fig:benchmark:2} }
\end{figure}

The experimental results are shown in \cref{fig:benchmark:1} and
\cref{fig:benchmark:2}. The lines summarize the statistics of utility gap metrics.
Both the mean and standard variance are presented in the figures. For the
clarity of presentation, we only draw the most three typical lines of UCB. From the two
figures, we can conclude that, within all the comparison algorithms, DCBO can
return the best parametric combination within a prescribed evaluation budget.
Moreover, its negative log likelihood decreases more rapidly than its
comparison algorithms. This means that its advantage appears not only in the end of
execution, but also arises over most iterations. The comparison on standard
deviations shows that in a majority of cases, DCBO performs well and can lead to a
consistent decrease. This observation agrees with our primal goal in setting up
the directional constraints.

\section{Conclusion}
\label{sec:conclusion}

In conclusion, there are many practical scenarios that standard Bayesian algorithm is 
ill-suited for. This is because
it does not find a way to cast the realistic constraints like evaluation budget into
the acquisition function in order to deal with the trade-off between speedup and the
gradually tighten ambient budget. 

An optimizer needs to be aware of the remaining evaluation budget. To do so, we
generalize the standard Bayesian optimization and propose a multiplicatively
reweighed searching policies, combining both global and local strategies. On the
local scale, we use a directional constraint to assist the optimizer, which can
be easily adapted into other probabilistic Bayesian optimization frameworks as a
plug-in module. This strategy performs well in reducing the jiggles exhibited in standard
expected improvement. The studies in real-world applications demonstrate that
this method could have substantial speedup without violating the prescribed
evaluation limitation.

Hitherto, we discuss the possibility of conducting Bayesian optimization in a
scenario with limited evaluation budget. However, there are prior knowledges that
are hard to be described in terms of probabilities. For instance, the partial
order existing among parameters and constraints on the parametric magnitude. In
addition, there exist a power in our formulation of the new acquisition function,
which controls the relative importance between two factors. This power is
considered to be a nuisance parameter. It is desirable for an adaptive function
that can automatically tune the balance between two factors. Both problems
need more complicated techniques, which constitute our future work.

\section*{Appendix}

The directional constraint for any given $ x $ takes a probabilistic form, with
the direction ending at $ x $ as its argument. From the Bayesian formalism, we
need to infer its posterior probability under an acute angle restriction. The
update rule for the posterior probability takes the following form:
\begin{align}
\label{eq:posterior}
H_{t+1}(x)& = p(g_{t+1}) \\
&= \int d g_t^\star d g_t \I(\langle g_{t+1}, g_t\rangle \ge 0) p(g_t) p(g_{t+1}|g^\star_t)p(g^\star_t) \notag
\end{align}
where $ g_{t+1} = \frac{x - x_t}{||x - x_t||} $.

To solve \cref{eq:posterior}, we need to infer $ p(g^\star) $ from the
GP which takes the form of a Gaussian distribution. This transformation extracts
directional uncertainties from the statistics of $ x^\star $. Suppose we have
the posterior estimation at the position $ x^\star $:
\begin{equation}
p(x^\star) = \mathcal{N}(\mu_{x^\star},\Sigma_{x^\star})
\end{equation}

Define a series of $ N $ samples $ \mathbf{x}_i $ draw from Gaussian
distribution $ \mathbf{x}_i, i={1,\cdots,N} $. Following \cite{Jammalamadaka2001,Banerjee2005}, we know immediately
that the maximum likelihood estimation of $ \theta $ and $ \kappa $ are give by
\begin{align}
\left\{
\begin{aligned}
\mathbf{\theta} &= \frac{\sum_{i=1}^{N} \mathbf{x}_i}{||\sum_{i = 1}^{N}\mathbf{x}_i||}\\
\kappa &= A_d^{-1}(\frac{1}{N}||\sum_{i=1}^{N} \mathbf{x}_i||)
\end{aligned}
\right.
\end{align}
where $ A_d = \frac{B_{d/2}(\kappa)}{B{d/2-1}(\kappa)} $.

Introducing an intermediate variable $ R =\frac{1}{N}||\sum_{i=1}^{N} x_i||  $, \cite{Banerjee2005} points out a
simple approximation to $ \kappa $ by
\begin{align}
\kappa = \frac{{R}(d - {R}^2)}{1 - {R}^2}
\end{align}

Suppose we adopt an Euclidean norm, the estimation of $ R $ could be easily
obtained through samples from the Gaussian distribution $ p(x^\star) $. 
\begin{align}
\label{eq:para:r}
\hat{R} &= \frac{1}{N}\E\sqrt{(\sum_{i=1}^N \frac{\mathbf{x}_i}{||\mathbf{x}_i||})^T(\sum_{i=1}^N \frac{\mathbf{x}_j}{||\mathbf{x}_i|| })}
\end{align}

Notationally, a random variable with hat represents the one estimated through
samples from a distribution. In summary, the key parameters in $
p(g^\star) $ are listed as follows:
\begin{align}
\left\{
\begin{aligned}
\hat{\theta} &= \frac{\mu}{\hat{R}} \\
\hat{\kappa} &= \frac{\hat{R}(d - \hat{R}^2)}{1 - \hat{R}^2}
\end{aligned}
\right.
\end{align}

Having obtained $ p(g^\star) $, the next step is to compute the
integral. Since there is no closed-form solution, the posterior distribution
needs to be carefully approximated. Provided that this posterior distribution is
in the same family as the prior probability, such that the chain rule of
Bayesian formalism could work. 

The posterior probability of $ g_{t+1} $ is formulated as:
\begin{align*}
p(g_{t+1}|x_t,g^\star) &= Vmf(g_{t+1}|\theta_{t+1},\kappa_{k+1})
\end{align*}
where $ \theta_{t+1} = g^\star =  \frac{\mu_{x^\star} - x_{t}}{||\mu_{x^\star} - x_{t}||} $.

From \cref{eq:posterior}, we know that the posterior probability can be split into a product of two independent terms:
\begin{align*}
&	p(g_{t+1}) \\ 
& = \int d g^\star d g_t \I(\langle g_{t+1}, g_t\rangle \ge 0) p(g_{t+1}|g^\star_t)p(g^\star_t) \\
& = \int d g^\star p(g_{t+1}|g^\star_t)p(g^\star_t) \int d g_t \I(\langle g_{t+1}, g_t\rangle \ge 0) p(g_t)
\end{align*}

Let us first concentrate on the first term in the above product.
\begin{align*}
\begin{split}
&\int d g^\star p(g_{t+1}|g^\star_t)p(g^\star_t) \\
&= \int_{||g^\star||=1 } d g^\star p(g_{t}|g^\star,x_{t})p(g^\star|x_{t}) \\
& = \int_{||g^\star||=1} d g^\star Vmf(g_{t+1}|g^\star, \kappa_{t})Vmf(g^\star|\hat{\theta}^\star,\hat{\kappa}^\star) \\
& = C_d(\kappa_{t})C_d(\hat{\kappa}^\star) \int_{ ||g^\star||=1} \exp(\kappa_{t} g^\star {}^T g_{t+1} + \hat{\kappa}^\star g^\star{}^T \hat{\theta}^{\star})\\
& = C_d(\kappa_{t})C_d(\hat{\kappa}^\star) \int_{||g^\star||=1} \exp\big((\kappa_{t}g_{t+1} + \hat{\kappa}^\star \hat{\theta}^\star)^T g^\star\big)
\end{split}
\end{align*}
where we include stationary points for ease of understanding. 

We use the fact that $ g_t $ is distributed according to $ Vmf $ of center $ \theta_t $ and concentration parameter $ \kappa_t $. The second term in the product can be computed as:
\begin{align*}
& \E\I(\langle g_{t+1}, g_t\rangle \ge 0) \\
=& \int_{g_{t+1} - \frac{\pi}{2}}^{g_{t+1} + \frac{\pi}{2}} Vmf(g_t|\theta_t,\kappa_t) d g_t\\
= & C \exp\big(\kappa_t \theta_t^T g_{t+1}\big)
\end{align*}
where the constant term $ C $ summarizes all irrelevant terms in terms of $
g_{t+1} $. Here and subsequently, we write $ C $ to stand for the constant that are
irrelevant to the variable of interest.

From the equality
\begin{equation*}
\int_{||\mathbf{x}||_2=1}\exp(\kappa\mathbf{\theta}^T\mathbf{x})\,d\mathbf{x} = \frac{(2\pi)^{d/2-1} B_{d/2-1}(\kappa)}{\kappa^{d/2-1}} \equiv C_d(\kappa)
\end{equation*}
it may be concluded that
\begin{align*}
\begin{split}
&p(g_{t+1}|x_t)\\
=& \E\I(\langle g_{t+1}, g_t\rangle \ge 0) C_d(\kappa_t) C_d(\hat{\kappa}^\star)C_d(|| \kappa_{t}g_{t+1} + \hat{\kappa}^\star \hat{\theta}^\star ||)\\
= & C \exp\big(\kappa_t \theta_t^T g_{t+1}\big)C_d(|| \kappa_{t}g_{t+1} + \hat{\kappa}^\star \hat{\theta}^\star ||)
\end{split}
\end{align*}
As defined above, $ C $ is a generic constant term that varies in the context.

If we hope to build a hierarchy model as Bayesian optimization does, it is
necessary to provide the posterior estimation for the $ \kappa $. A well-known
equality for distribution in exponential family states that:
\begin{equation}
\label{eq:ktheta}
\frac{\partial \log p(x)}{\partial x} = \kappa \mathbf{\theta}
\end{equation}

Then, the parameter $ \kappa_{t+1} $ are approximated
with the following equations.
\begin{align*}
&\frac{\partial}{\partial g_{t+1}} \log p(g_{t+1}) \\
&= \frac{\partial }{\partial g_{t+1}}\log \big( C\exp(\kappa_t \theta_t^T g_{t+1})C_d(|| \kappa_{t}g_{t+1} + \hat{\kappa}^\star \hat{\theta}^\star ||) \big)\\
& = \frac{\partial}{\partial g_{t+1}} \log C_d(|| \kappa_{t}g_{t+1} + \hat{\kappa}^\star \hat{\theta}^\star ||) + k_t \theta_t
\end{align*}

This is not a properly defined statistical quantity, but we approximate this quantity with probability in the same family. After defining an intermediate variable $ y = || \kappa_{t}g_{t+1} + \hat{\kappa}^\star \hat{\theta}^\star || $, we approximate it with the Taylor series in terms of $ g_{t+1} $, which can be obtained as:
\begin{align}
\label{eq:obj:y}
y = &\sqrt{\hat{\kappa}^\star {}^2 + \kappa_t^2 } + \frac{\hat{\kappa}^\star \kappa_t\hat{\theta}^\star g_{t+1}}{\sqrt{\hat{\kappa}^\star {}^2 + \kappa_t^2 }}
+ \frac{(\hat{\kappa}^\star \kappa_t\hat{\theta}^\star)^2 g_{t+1}^2}{2(\hat{\kappa}^\star {}^2 + \kappa_t^2 )^{3/2}} + O(g_{t+1}^3)
\end{align}

Taking derivative on both sides of \cref{eq:obj:y} with respect to $ g_{t+1} $, we get 
\begin{align}
\label{eq:div:1}
\frac{d y}{d g_{t+1}} = \frac{\hat{\kappa}^\star \kappa_t\hat{\theta}^\star}{\sqrt{\hat{\kappa}^\star {}^2 + \kappa_t^2 }} + \frac{(\hat{\kappa}^\star \kappa_t\hat{\theta}^\star)^2 g_{t+1}}{(\hat{\kappa}^\star {}^2 + \kappa_t^2 )^{3/2}} + O(g_{t+1}^2)
\end{align}

We notice that the function $ \log C_d(y) $ could be decomposed as follows: 
\begin{align}
\label{eq:div}
\begin{split}
& \log C_d(y) \\
&\quad = log \frac{(2\pi )^{d/2-1}B_{d/2-1}(y)}{y^{d/2-1}}\\
& \quad = (\frac{d}{2}-1) \log{2\pi} + \log B_{\frac{d}{2}-1}(y) - (\frac{d}{2}-1)\log{y}
\end{split}
\end{align}

Again, taking derivative on both sides of \cref{eq:div}, we get 
\begin{align}
\label{eq:div:2}
\begin{split}
\frac{d}{dy}\log C_{d}(y)
&= - (\frac{d}{2}-1)\frac{1}{y} + \frac{d}{dy} \log B_{\frac{d}{2}-1}(y)\\
& \approx \frac{y}{d}-\frac{y^3}{d^2 (d+2)}+O\left(y^4\right)
\end{split}
\end{align}
during which, we make use of an equality
\[\frac{d}{dy} \log B_{\frac{d}{2}-1}(y) =\frac{B_{\frac{d}{2}}(y)+B_{\frac{d}{2}-2}(y)}{2 B_{\frac{d}{2}-1}(y)}\] and expand the results in a Taylor series.

Put \cref{eq:div:1} and \cref{eq:div:2} together, we get 
\begin{align*}
& \frac{d}{dy}C_d(y)\frac{dy}{d g_{t+1}}\\
&= -\frac{\hat{\kappa}^\star \kappa_{t} \hat{\theta}^\star \left(-d (d+2)+\hat{\kappa}^\star {}^2+\kappa_{t}^2\right)}{d^2 (d+2)}+ O\left(g_{t+1}\right) \\
&\approx -\frac{\hat{\kappa}^\star \kappa_{t} \hat{\theta}^\star \left(-d (d+2)+\hat{\kappa}^\star {}^2+\kappa_{t}^2\right)}{d^2 (d+2)}\\
& \triangleq k_1 \hat{\theta}^\star\\
\end{align*}
where the third equality uses the fact that $ ||\hat{\theta}^\star|| = 1 $. In the fourth equation, we introduce two intermediate variables:
\begin{align*}
k_1 &= -\frac{\hat{\kappa}^\star \kappa_{t} \left(-d (d+2)+\hat{\kappa}^\star {}^2+\kappa_{t}^2\right)}{d^2 (d+2)}\\
\end{align*}

By the fact that $ ||\theta_{t+1}|| = 1 $, we reach the final update rules:
\begin{align}
\label{eq:est}
\left\{
\begin{aligned}
\kappa_{t+1} &= \sqrt{k_1^2 + \kappa_t^2} \\
\theta_{t+1} &= \frac{1}{\kappa_{t+1}}k_1 \hat{\theta}^\star + \frac{1}{\kappa_{t+1}}\kappa_t\theta_t
\end{aligned}
\right.
\end{align}


\bibliographystyle{IEEEtran}
\bibliography{bibtex}

\begin{thebibliography}{10}
\providecommand{\url}[1]{#1}
\csname url@samestyle\endcsname
\providecommand{\newblock}{\relax}
\providecommand{\bibinfo}[2]{#2}
\providecommand{\BIBentrySTDinterwordspacing}{\spaceskip=0pt\relax}
\providecommand{\BIBentryALTinterwordstretchfactor}{4}
\providecommand{\BIBentryALTinterwordspacing}{\spaceskip=\fontdimen2\font plus
\BIBentryALTinterwordstretchfactor\fontdimen3\font minus
  \fontdimen4\font\relax}
\providecommand{\BIBforeignlanguage}[2]{{%
\expandafter\ifx\csname l@#1\endcsname\relax
\typeout{** WARNING: IEEEtran.bst: No hyphenation pattern has been}%
\typeout{** loaded for the language `#1'. Using the pattern for}%
\typeout{** the default language instead.}%
\else
\language=\csname l@#1\endcsname
\fi
#2}}
\providecommand{\BIBdecl}{\relax}
\BIBdecl

\bibitem{Mockus1975}
J.~Mo{\v{c}}kus, ``On bayesian methods for seeking the extremum,'' in
  \emph{Optimization Techniques IFIP Technical Conference}.\hskip 1em plus
  0.5em minus 0.4em\relax Springer, 1975, pp. 400--404.

\bibitem{Jones2001}
D.~R. Jones, ``A taxonomy of global optimization methods based on response
  surfaces,'' \emph{Journal of global optimization}, vol.~21, no.~4, pp.
  345--383, 2001.

\bibitem{Srinivas2010}
N.~Srinivas, A.~Krause, S.~Kakade, and M.~Seeger, ``Gaussian process
  optimization in the bandit setting: No regret and experimental design,'' in
  \emph{International Conference on International Conference on Machine
  Learning}, 2010, pp. 1015--1022.

\bibitem{Kushner1964}
H.~J. Kushner, ``A new method of locating the maximum point of an arbitrary
  multipeak curve in the presence of noise,'' \emph{Journal of Basic
  Engineering}, vol.~86, no.~1, pp. 97--106, 1964.

\bibitem{Srinivas2012}
N.~Srinivas, A.~Krause, S.~M. Kakade, and M.~W. Seeger,
  ``Information-{Theoretic} {Regret} {Bounds} for {Gaussian} {Process}
  {Optimization} in the {Bandit} {Setting},'' \emph{IEEE Transactions on
  Information Theory}, vol.~58, no.~5, pp. 3250--3265, May 2012.

\bibitem{Bull2011}
A.~D. Bull, ``Convergence rates of efficient global optimization algorithms,''
  \emph{Journal of Machine Learning Research}, vol.~12, no. Oct, pp.
  2879--2904, 2011.

\bibitem{Hennig2012}
P.~Hennig and C.~J. Schuler, ``Entropy search for information-efficient global
  optimization,'' \emph{Journal of Machine Learning Research}, vol.~13, no.
  Jun, pp. 1809--1837, 2012.

\bibitem{Hernandez-Lobato2015}
J.~M. Hern{\'{a}}ndez-Lobato, M.~A. Gelbart, M.~W. Hoffman, R.~P. Adams, and
  Z.~Ghahramani, ``Predictive {Entropy} {Search} for {Bayesian} {Optimization}
  with {Unknown} {Constraints},'' in \emph{International Conference on Machine
  Learning}, vol.~37, 2015, pp. 1699--1707.

\bibitem{hernandez-lobato_predictive_2016}
D.~Hern{\'{a}}ndez-Lobato, J.~M. Hern{\'{a}}ndez-Lobato, A.~Shah, and R.~P.
  Adams, ``Predictive {Entropy} {Search} for {Multi}-objective {Bayesian}
  {Optimization},'' in \emph{{International} {Conference} on {Machine}
  {Learning}}, vol.~48, 2016, pp. 1492--1501.

\bibitem{Bishop2006}
C.~M. Bishop, \emph{Pattern {Recognition} and {Machine} {Learning}}.\hskip 1em
  plus 0.5em minus 0.4em\relax Springer, 2006, vol.~4, no.~4.

\bibitem{Rasmussen2006}
C.~E. Rasmussen and C.~K.~I. Williams, \emph{Gaussian processes for machine
  learning}, ser. Adaptive computation and machine learning.\hskip 1em plus
  0.5em minus 0.4em\relax Cambridge, Mass: MIT Press, 2006.

\bibitem{Brochu2010}
E.~Brochu, V.~M. Cora, and N.~de~Freitas, ``A {Tutorial} on {Bayesian}
  {Optimization} of {Expensive} {Cost} {Functions}, with {Application} to
  {Active} {User} {Modeling} and {Hierarchical} {Reinforcement} {Learning},''
  \emph{arXiv:1012.2599 [cs]}, Dec. 2010.

\bibitem{Jones1998}
D.~R. Jones, M.~Schonlau, and W.~J. Welch, ``\BIBforeignlanguage{en}{Efficient
  {Global} {Optimization} of {Expensive} {Black}-{Box} {Functions}},''
  \emph{\BIBforeignlanguage{en}{Journal of Global Optimization}}, vol.~13,
  no.~4, pp. 455--492, Dec. 1998.

\bibitem{Gardner2014}
J.~R. Gardner, M.~J. Kusner, Z.~E. Xu, K.~Q. Weinberger, and J.~P. Cunningham,
  ``Bayesian {Optimization} with {Inequality} {Constraints}.'' in
  \emph{International {Conference} on {Machine} {Learning}}, 2014, pp.
  937--945.

\bibitem{Picheny2016}
V.~Picheny, R.~B. Gramacy, S.~M. Wild, and S.~L. Digabel, ``Bayesian
  optimization under mixed constraints with a slack-variable augmented
  {Lagrangian},'' in \emph{Advances in {Neural} {Information} {Processing}
  {Systems}}, 2016, pp. 1435--1443.

\bibitem{Bernardo2011}
J.~Bernardo, M.~J. Bayarri, J.~O. Berger, A.~P. Dawid, D.~Heckerman, A.~F.~M.
  Smith, and M.~West, ``Optimization under unknown constraints,''
  \emph{Bayesian Statistics}, vol.~9, no.~9, p. 229, 2011.

\bibitem{Picheny2014}
V.~Picheny, ``\BIBforeignlanguage{en}{A {Stepwise} uncertainty reduction
  approach to constrained global optimization},'' in
  \emph{\BIBforeignlanguage{en}{Artificial {Intelligence} and {Statistics}}},
  Apr. 2014, pp. 787--795.

\bibitem{Lam2017}
R.~Lam and K.~Willcox, ``Lookahead {Bayesian} {Optimization} with {Inequality}
  {Constraints},'' in \emph{Advances in {Neural} {Information} {Processing}
  {Systems}}, I.~Guyon, U.~v. Luxburg, S.~Bengio, H.~M. Wallach, R.~Fergus,
  S.~V.~N. Vishwanathan, and R.~Garnett, Eds., 2017, pp. 1888--1898.

\bibitem{Lam2016}
R.~Lam, K.~Willcox, and D.~H. Wolpert, ``Bayesian optimization with a finite
  budget: An approximate dynamic programming approach,'' in \emph{Advances in
  Neural Information Processing Systems}, 2016, pp. 883--891.

\bibitem{Jammalamadaka2001}
S.~R. Jammalamadaka and A.~Sengupta, \emph{Topics in circular statistics}, ser.
  Series on multivariate analysis.\hskip 1em plus 0.5em minus 0.4em\relax River
  Edge, N.J: World Scientific, 2001, no. v. 5.

\bibitem{Banerjee2005}
A.~Banerjee, I.~S. Dhillon, J.~Ghosh, and S.~Sra, ``Clustering on the {Unit}
  {Hypersphere} {Using} {Von} {Mises}-{Fisher} {Distributions},'' \emph{J.
  Mach. Learn. Res.}, vol.~6, pp. 1345--1382, Dec. 2005.

\bibitem{Martens2010}
J.~Martens, ``Learning the linear dynamical system with {ASOS},'' in
  \emph{{International} {Conference} on {Machine} {Learning}}, 2010, pp.
  743--750.

\bibitem{Taylor2007}
G.~W. Taylor, G.~E. Hinton, and S.~T. Roweis, ``Modeling human motion using
  binary latent variables,'' in \emph{Advances in neural information processing
  systems}, 2007, pp. 1345--1352.

\end{thebibliography}
\end{document}